\documentclass[10pt,twocolumn,letterpaper]{article}

\usepackage{iccv}
\usepackage{times}
\usepackage{epsfig}
\usepackage{graphicx}
\usepackage{amsmath}
\usepackage{amssymb}
\usepackage{multirow}
\usepackage{makecell}
\usepackage{booktabs}
\usepackage[tracking=true]{microtype}
\usepackage{flushend}

\usepackage[pagebackref=true,breaklinks=true,colorlinks,bookmarks=false]{hyperref}
\usepackage[table,xcdraw]{xcolor}
\def\iccvPaperID{3244} 

\newcommand{\MethodName}{PatchMatch-RL}

\iccvfinalcopy 

\ificcvfinal\fi

\begin{document}

\title{
PatchMatch-RL: Deep MVS with Pixelwise Depth, Normal, and Visibility}

\author{Jae Yong Lee$^1$\\
\and
Joseph DeGol$^2$\\
\and
Chuhang Zou\thanks{Work done outside of Amazon}\hspace{2pt} $^3$\\
\and
Derek Hoiem$^1$
\and
$^1$\textls[-20]{University of Illinois at Urbana-Champaign}\\
\textls[-30]{\tt\small \{lee896, dhoiem\}@illinois.edu}
\and
$^2$Microsoft\\
\textls[-30]{\tt\small joseph.degol@microsoft.com}
\and
$^3$Amazon Go\\
\textls[-30]{\tt\small zouchuha@amazon.com}
}

\maketitle

\begin{abstract}
    Recent learning-based multi-view stereo (MVS) methods show excellent performance with dense cameras and small depth ranges. 
    However, non-learning based approaches still outperform for scenes with large depth ranges and sparser wide-baseline views, in part due to their PatchMatch optimization over pixelwise estimates of depth, normals, and visibility.  
    In this paper, we propose an end-to-end trainable PatchMatch-based MVS approach that combines advantages of trainable costs and regularizations with pixelwise estimates. To overcome the challenge of the non-differentiable PatchMatch optimization that involves iterative sampling and hard decisions, we use reinforcement learning to minimize expected photometric cost and maximize likelihood of ground truth depth and normals. We incorporate normal estimation by using dilated patch kernels and propose a recurrent cost regularization that applies beyond frontal plane-sweep algorithms to our pixelwise depth/normal estimates.
    We evaluate our method on widely used MVS benchmarks, ETH3D and Tanks and Temples (TnT). On ETH3D, our method outperforms other recent learning-based approaches and performs comparably on advanced TnT. 
\end{abstract}
\urlstyle{sf}
\let\thefootnote\relax\footnote{
Code available at \url{https://github.com/leejaeyong7/patchmatch-rl}
}
\urlstyle{tt}
\section{Introduction}
\begin{figure}[t]
    \centering
    \begin{tabular}{@{}c@{}c@{}}
    \includegraphics[width=0.235\textwidth]{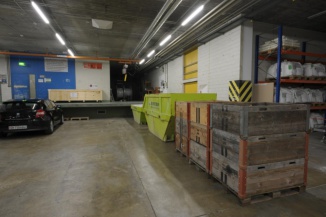} & 
    \includegraphics[width=0.235\textwidth]{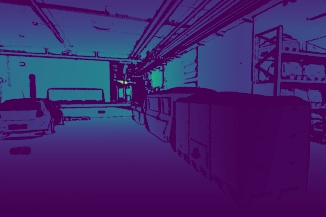} 
    \\
    Image & Ground Truth \\
    \includegraphics[width=0.235\textwidth]{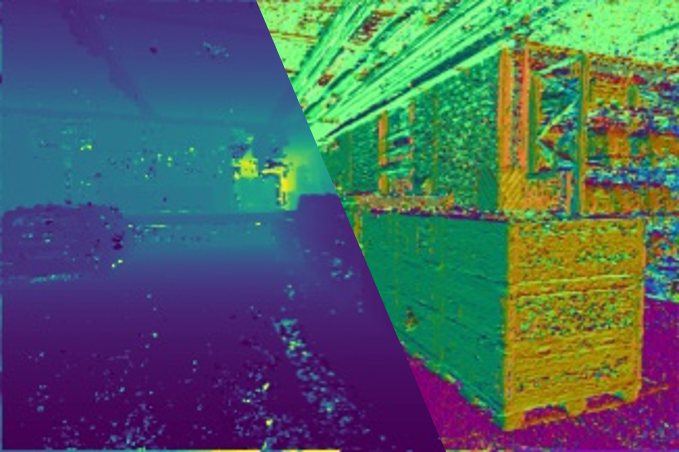} & 
    \includegraphics[width=0.235\textwidth]{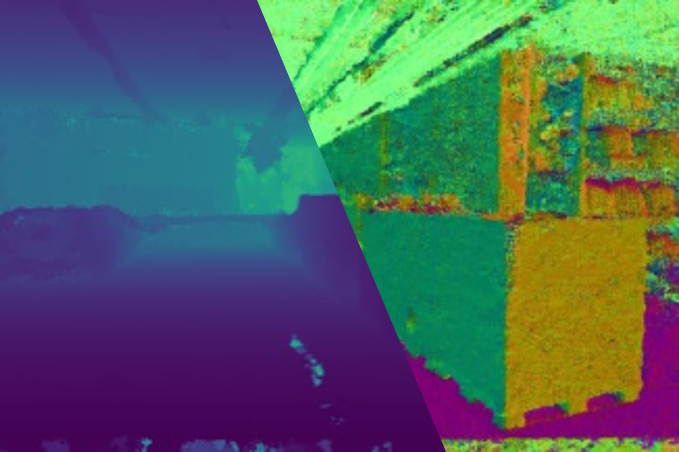} \\
    COLMAP~\cite{schoenberger2016mvs} & Ours \\
    \end{tabular}
    
    \caption{
        We propose \textbf{PatchMatch-RL}, an end-to-end trainable PatchMatch-based MVS approach that combines advantages of trainable costs and regularizations with pixelwise estimates of depth, normal, and visibility. The left half of the bottom images is the depth, and the right half is the normals. We show that our method can achieve smoother and more complete depth and normal map estimation over the existing approach (COLMAP).
    }
    \label{fig:teasure}
 \end{figure}

Multi-view stereo (MVS) aims to reconstruct 3D scene geometry from a set of RGB images with known camera poses, with many important applications such as robotics~\cite{rebecq2018emvs}, self-driving cars~\cite{duggal2019deeppruner}, infrastructure inspection~\cite{DeGol:CVPR:16,hanConstruction}, and mapping~\cite{wang2017stereo}.
Non-learning based MVS methods~\cite{bleyer2011patchmatch, romanoni2019tapa, xu2019multi, Xu2020ACMP, zheng2014patchmatch} evolved to support pixelwise estimates of depths, normals, and source view selection, with PatchMatch based iterative optimization and cross-image consistency checks. Recent learning-based MVS methods~\cite{Guo_2019_CVPR, huang2018deepmvs, ji2017surfacenet, yu2020fast, vzbontar2016stereo} tend to use frontal plane sweeps, evaluating the same set of depth candidates for each pixel based on the same images. The trainable photometric scores and cost-volume regularization of the learning-based methods leads to excellent performance with dense cameras and small depth ranges, as evidenced in the DTU~\cite{aanaes2016large} and Tanks-and-Temples (TnT) benchmarks~\cite{Knapitsch2017tanks}, but the pixelwise non-learning based approach outperforms for scenes with large depth ranges and slanted surfaces observed with sparser wide-baseline views, as evidenced in the ETH3D benchmark~\cite{schoeps2017eth3d}. 

Our paper aims to incorporate pixelwise depth, normal, and view estimates into an end-to-end trainable system with advantages from both approaches:
\begin{itemize}
    \vspace{-0.5em}
    \item \textbf{Pixelwise depth and normal} prediction efficiently models scenes with large depth ranges and slanted surfaces.
    \vspace{-0.5em}
    \item  \textbf{Pixelwise view selection} improves robustness to occlusion and enables reconstruction from sparser images.
    \vspace{-0.5em}
    \item \textbf{Learned photometric cost} functions improve correspondence robustness.
    \vspace{-0.5em}
    \item \textbf{Learned regularization and contextual inference} enable completion of textureless and glossy surfaces.
    \vspace{-1.0em}
\end{itemize}

One challenge is that PatchMatch optimization and pixelwise view selection involve iterative sampling and hard decisions that are not differentiable. We propose a reinforcement learning approach to minimize expected photometric cost and maximize discounted rewards for reaching a good final solution.  Our techniques can also be used to enable learning for other PatchMatch applications (e.g.~\cite{barnes2009patchmatch,hu2016efficient,lu2013patch}), though we focus on MVS only.  Estimating 3D normals of pixels is also challenging because convolutional features tend to be smooth so that neighboring cells add little new information, and patch-wise photometric costs are memory intensive. We find that with shallower feature channels and dilated patch kernels, we effectively estimate pixel normals. A third challenge is how to perform regularization or global inference. Each pixel has its own depth/normal estimate, so cost-volume based regularization does not apply.  We propose a recurrent cost regularization that updates a hidden state via message passing that accounts for depth/normal similarities between pixels.

\begin{figure*}[ht] \centering
    \includegraphics[width=0.98\textwidth]{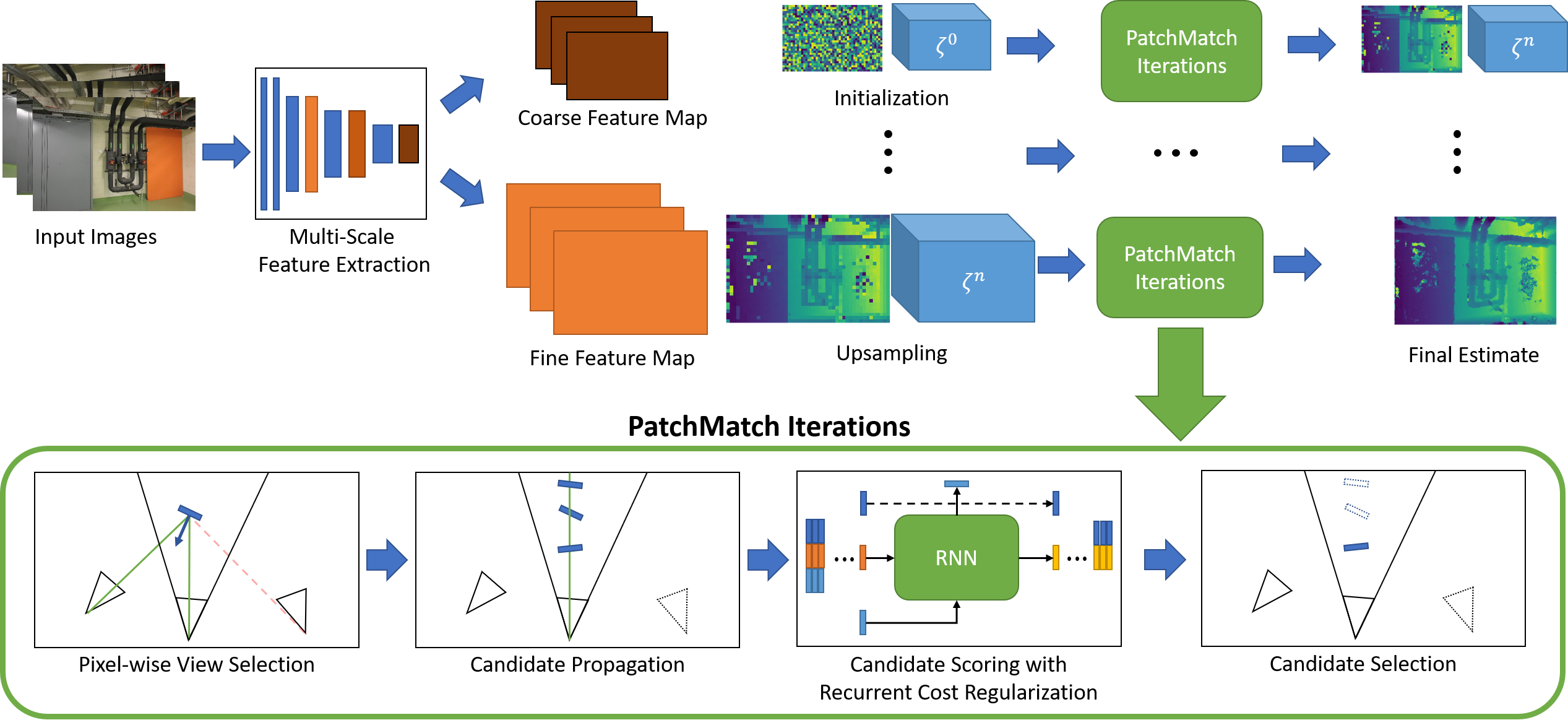}
    \caption{
        \textbf{Architecture overview: }
        We first extract multi-scale features using CNNs with shared weights. We then perform coarse-to-fine estimation, with correlation of feature maps at corresponding scales used to evaluate photometric costs and perform view selection.
        At the coarsest stage, we initialize pixelwise oriented points (depths/normals) and associated hidden states per plane. Then, a series of PatchMatch iterations updates the points and hidden state maps.
        The PatchMatch iteration consists of four stages: (1) pixelwise view selection; (2) candidate propagation; (3) candidate scoring with recurrent cost regularization; and (4) candidate selection.  The current solution is then upsampled as an input to the finer level, and this continues until oriented point estimates at the finest level are fused from all images.
    }
    \label{fig:architecture}
 \end{figure*}


In summary, our main \textbf{contribution} is an end-to-end trainable PatchMatch-based MVS approach that combines advantages of trainable costs and regularizations with pixelwise estimates, requiring multiple innovations:
\begin{itemize}
    \vspace{-0.5em}
    \item Reinforcement learning approach to train end-to-end within a PatchMatch sampling based optimization.
    \vspace{-0.5em}
    \item Use of normal estimates in learning-based MVS, enabled by trainable PatchMatch optimization and CNN patch features.
    \vspace{-0.5em}
    \item Depth/normal regularization that applies beyond frontal plane-sweep algorithms; e.g. to our pixelwise depth/normal estimates.
    \vspace{-0.5em}
\end{itemize}
In experiments, our system outperforms other recent learning-based methods on ETH3D and performs similarly on TnT, and our ablation study validates the importance  of pixelwise normal and view selection estimates.


\section{Related Works}

Given correct scene geometry, the pixels that correspond to a surface patch in different calibrated cameras can be determined, and their appearance patterns will be similar (``photometrically consistent'').  This core idea of multi-view stereo (MVS) leads to an array of formulations, optimization algorithms, and refinements. We focus on our work's direct lineage, referring the interested reader to a survey/tutorial~\cite{furukawa2015_mvsTutorial} and paper list~\cite{awesome_3d_list} for more complete background and coverage.

The first and simplest formulation is to assign each pixel to one of a set of candidate disparities or depth values~\cite{lucasKanade1981_stereo}. The locally best assignment can be determined by filtering across rows in rectified images, and surface smoothness priors can be easily incorporated within this ordered labeling problem.  However, per-view depth labeling has many shortcomings in a wide-baseline MVS setting: (1) depth maps do not align in different views, making consistency checking and fusion more difficult; (2) depth for oblique surfaces is not constant, degrading matching of intensity patches; and (3) the range of depth values may be large, so that large steps in depth are needed to feasibly evaluate the full range.  Further, occlusion and partially overlapping images demand more care in evaluating photometric consistency.  

These difficulties led to a reformulation of MVS as solving for a depth, normal, and view selection for each pixel in a reference image~\cite{schoenberger2016mvs, zheng2014patchmatch}.  The view selection identifies which other source images will be used to evaluate photometric consistency. This more complex formulation creates a challenging optimization problem, since each pixel has a 4D continuous value (depth/normal) and binary label vector (view selection). PatchMatch~\cite{barnes2009patchmatch, bleyer2011patchmatch, schoenberger2016mvs} is well-suited for the depth/normal optimization, since it employs a hypothesize-test-propagate framework that is ideal for efficient inference when labels have a large range but are approximately piecewise constant in local neighborhoods. The pixelwise PatchMatch formulations have been refined with better propagation schemes~\cite{xu2019multi},  multi-scale features~\cite{xu2019multi}, and plane priors~\cite{romanoni2019tapa,Xu2020ACMP}. Though this line of work addresses the shortcomings of the depth labeling approach, it often fails to reconstruct smooth or glossy surfaces where photometric consistency is uninformative, mainly due to the challenge of incorporating global priors, which is addressed in part by Kuhn et al.'s post-process trainable regularization~\cite{kuhn2020deepc}. Also, though hancrafted photometric consistency functions, such as bilaterally weighted NCC, perform well in general, learned functions can potentially outperform by being context-sensitive.

Naturally, the first inroads to fully trainable MVS also followed the simplistic depth labeling formulation~\cite{huang2018deepmvs,ji2017surfacenet,yao2018mvsnet}, which comfortably fits the CNN forte  of learning features, performing inference over ``cost volumes'' (features or scores for each position/label), and producing label maps.  But despite improvements such as using recurrent networks~\cite{yao2019recurrent} to refine estimates, coarse-to-fine reconstruction~\cite{yu2020fast}, visibility maps~\cite{xu2020pvsnet}, and attention-based regularization~\cite{luo2020attention}, many of the original drawbacks of the depth labeling formulation persist.   

Thus, we now have two parallel branches of MVS state-of-the-art: (1) complex hand-engineered formulations with PatchMatch optimization that outperform for large-scale scene reconstruction from sparse wide-baseline views; and (2) deep network depth-labeling formulations that outperform for smaller scenes, smooth surfaces, and denser views.  Differentiation-based learning and sampling-based optimization are not easily reconciled with refinements or combinations of existing approaches. Duggal et al.~\cite{duggal2019deeppruner} propose a differentiable PatchMatch that optimizes softmax-weighted samples, instead of argmax, and use it to prune the depth search space to initialize depth labeling. We use their idea of one-hot filter banks to perform propagation but use an expectation based loss that sharpens towards argmax during training to enable argmax inference. The very recent PatchmatchNet~\cite{wang2020patchmatchnet} minimizes a sum of per-iteration losses and employs a one-time prediction of visibility (soft view selection).  We use reinforcement learning to train view selection and minimize the loss of the final depth/normal estimates.
Our work is the first, to our knowledge, to propose an end-to-end trainable formulation that combines the advantages of pixelwise depth/normal/view estimates and PatchMatch optimization with deep network learned photometric consistency and refinement. 
\section{PatchMatch-RL MVS}
We propose \MethodName, an end-to-end learning framework that uses PatchMatch for Multi-View Stereo (MVS) reconstruction. 
Figure~\ref{fig:architecture} shows an overview of our approach. 
Given a set of images $\mathcal{I}$ and its corresponding camera poses $\mathcal{C} = (\mathcal{K}, \mathcal{E})$ with intrinsic $\mathcal{K}$ and extrinsic $\mathcal{E}=[\mathcal{R}, t]$ matrices, our goal is to recover the depths (and normals) of the reference image $I_{ref}\in \mathcal{I}$ using a set of selected source images $I_{src} \subset \mathcal{I}$ that overlap with $I_{ref}$. 

Rather than solving only for depth, we also estimate surface normals, which enables propagating hypotheses and comparing spatially distributed features between reference and source images along the local plane.  Surface normal estimation improves depth estimates for oblique surfaces and is also useful for consistency checks, surface modeling, and other downstream processing. 

Our estimation proceeds coarse-to-fine. At the coarsest level, estimates are randomly initialized and then refined through a series of PatchMatch iterations that consist of pixelwise view selection, candidate propagation, regularized cost computation, and candidate update. Resulting estimates are then upsampled and further refined, this continues until the finest layer, after which all depth estimates are fused into a 3D point cloud.  

\subsection{Initialization} 

For each level in the coarse-to-fine optimization, we extract CNN features for the reference and source images using a Feature Pyramid Network (FPN)~\cite{lin2017feature}. For memory efficiency, the number of output channels varies per scale, with shallower feature channels in the higher-resolution feature maps. $\mathcal{F}^s_p$ denotes the feature vector for pixel $p$ at image $s$. 

Our goal is to solve for an oriented point $\omega_p$, consisting of a plane-camera distance $\delta_p$ and normal $\textbf{n}_p$, for each pixel $p$ in $I_{ref}$. Pixel depth $d_p$ is related to $\delta_p$ through $d_p = -\delta_p / (\mathbf{n}_p \cdot \mathcal{K}^{-1}\cdot p)$. 
The depth $d_p$ is sampled uniformly from the inverse depth range as: $d^0_p \sim 1/\mathcal{U}(\frac{1}{d_{max}}, \frac{1}{d_{min}})$, with $d_{min}$ and $d_{max}$ specifying the depth range. 
Sampling from the inverted range prioritizes depths closer to the camera center, as shown effective by Gallup \etal~\cite{galliani2015massively}. 
The per-pixel normal $\mathbf{n}_p$ is initialized independently of depth by sampling from a 3D Gaussian and applying L2 normalization~\cite{muller1959note}. 
The normal vector is reversed if it faces the same direction as the pixel ray. 

\subsection{Feature Correlation}

The feature maps can be differentiably warped~\cite{yao2018mvsnet} according to the pixelwise plane homographies from reference image $r$ to source image $s$ as $H^{r\rightarrow s}_{\omega_p} = \mathcal{K}_s\cdot (\mathcal{R}_{r \rightarrow s} - \frac{t_{r\rightarrow s}\textbf{n}_p^{T}}{\delta_p}) \cdot \mathcal{K}_r^{-1}$. 
With support window $\mathcal{W}^{\alpha, \beta}_p$ of size $\alpha$ and dilation $\beta$ centered at $p$, we define the  correlation value $\mathcal{G}^s_{\omega_p}$ of the oriented point $\omega_p$ as the attention-aggregated group-wise correlation for matching feature vectors in the source image:
\begin{align*}
   \mathcal{A}^q_p &= \sigma(\frac{\mathcal{F}^{r}_p \cdot h}{\sqrt{||\mathcal{F}^{r}_p||_2}})_q, q\in \mathcal{W}^{\alpha, \beta}_p \\
   \mathcal{G}^s_{\omega_p} &= \sum_{q} \mathcal{A}^q_p \cdot (\mathcal{F}^{r}_q \circledast \mathcal{F}^s_{H^{r \rightarrow s}_{\omega_p} \cdot q}).
\end{align*}
We denote group-wise feature vector correlation~\cite{xu2020learning} as $\circledast$, scaled dot-product attention for supporting pixel $q$ on center pixel $p$ by the reference feature map as $\mathcal{A}^q_p$, and the attentional feature projection vector as $h$, implemented as a 1x1 convolution.
The resulting $\mathcal{G}^s_{\omega_p}$ represents the similarity of the features centered at $p$ in the reference image and the corresponding features in the source image, according to $\omega_p$. 

In preliminary experiments, our estimation of normals $\mathbf{n}_p$ was poor and did not improve depth estimation. The problem was that the smoothness of features prevented a 3x3 patch from providing much additional information. Making larger patches was not practical due to memory constraints. This problem was solved through use of dilation ($\beta=3$), and we further reduced memory usage by producing shallower feature channels. 


\begin{figure}[t]
    \centering
    \includegraphics[width=0.152\textwidth]{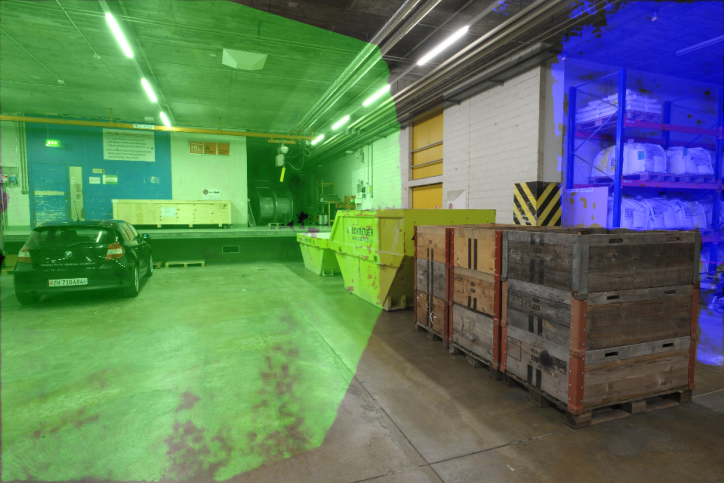} 
    \includegraphics[width=0.152\textwidth]{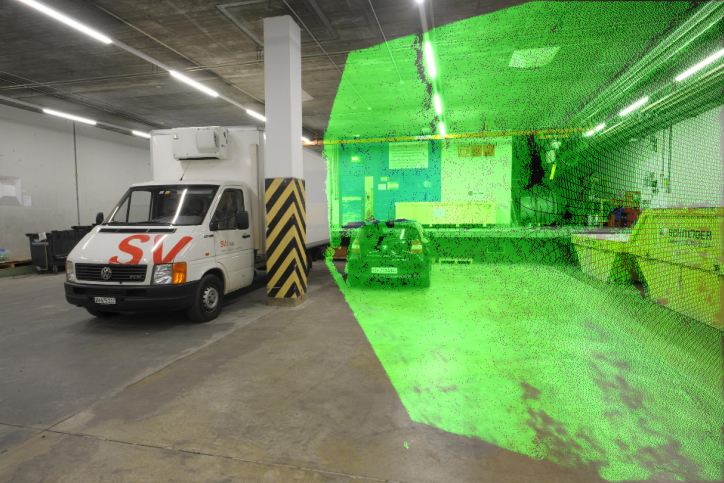}
    \includegraphics[width=0.152\textwidth]{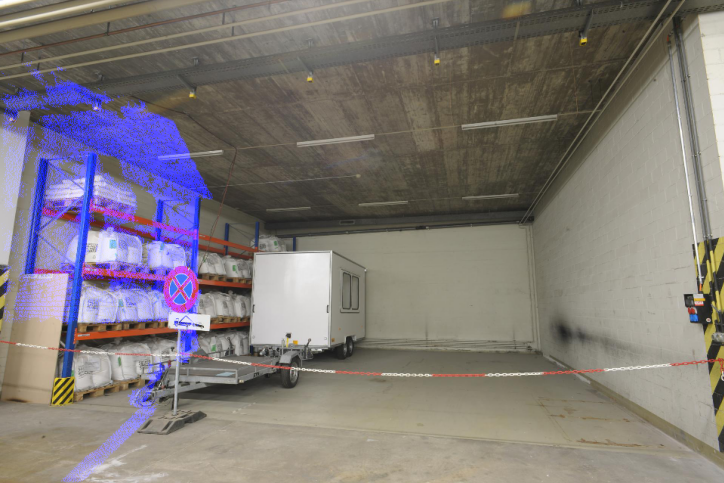}
    \caption{
        \textbf{Estimated Visibilities.}
        The leftmost image corresponds to the reference image, and the right two images are the source images. 
        We color-code the corresponding region to be estimated as visible in the last PatchMatch iteration. The estimated visibility matches precisely with the actual visibility. 
        (Best viewed in color.)
    }
     
    \label{fig:view-selection}
 \end{figure}

\subsection{Pixel-wise View Selection}
\label{sec:pixelwise_view_selection}

Based on Schönberger et al.~\cite{schoenberger2016mvs}, we compute scale, incident-angle, and triangulation angle difference based geometric priors for each source image $s$ for each $\omega_p$. Instead of hand-crafting the prior function, we concatenate the priors with the feature correlations $\mathcal{G}^{s}_{\omega_p}$ and use a multi-layered perceptron (MLP) to predict a pixel-wise visibility estimate, denoted $\hat{\mathcal{V}}^s_p \in [0, 1]$.
Figure~\ref{fig:view-selection} shows an example of the estimated visibilities in the source images.

We then sample $N$-views based on the L1 normalized probability distribution over $\hat{\mathcal{V}}^s_p$ for each pixel, to obtain a sampled set of views, $\mathcal{V}_p$. The visibility probabilities are further used to compute a weighted sum of feature correlations across views. 




\subsection{Candidate Propagation}
\begin{figure}[t]
    \centering
    \begin{tabular}{ccc}
        \includegraphics[width=0.135\textwidth]{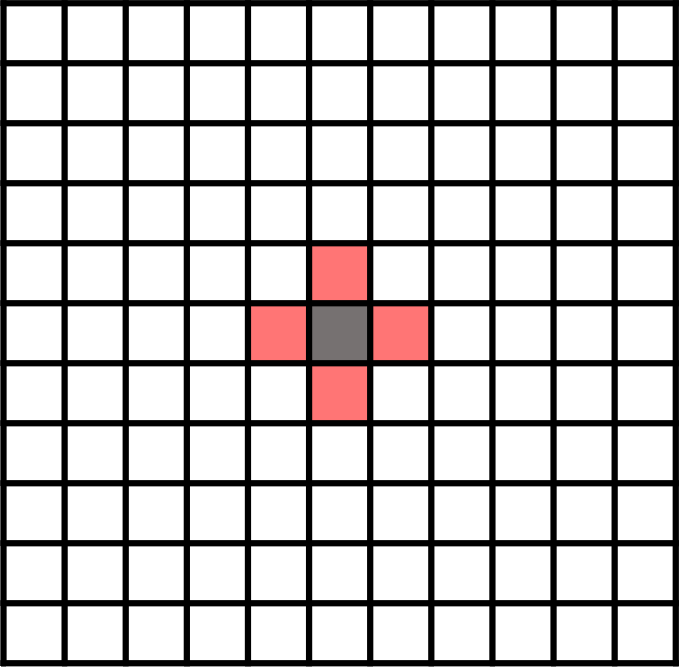}&
        \includegraphics[width=0.135\textwidth]{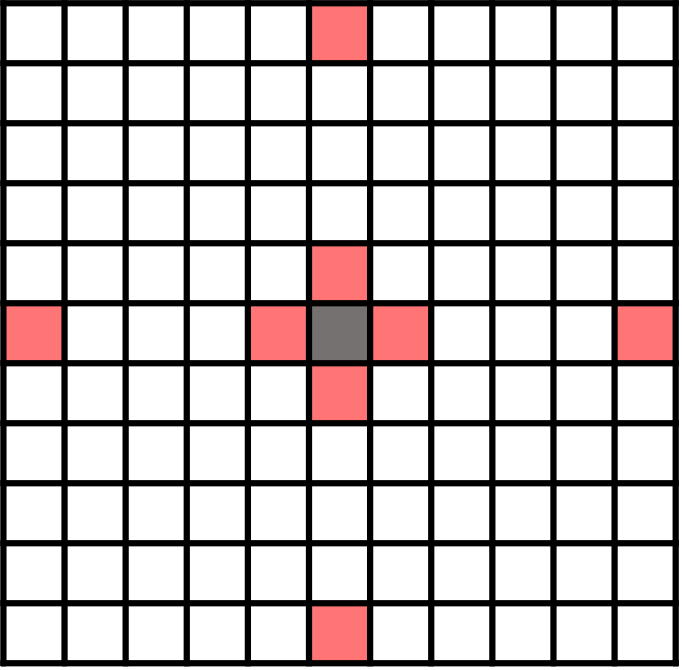}&
        \includegraphics[width=0.135\textwidth]{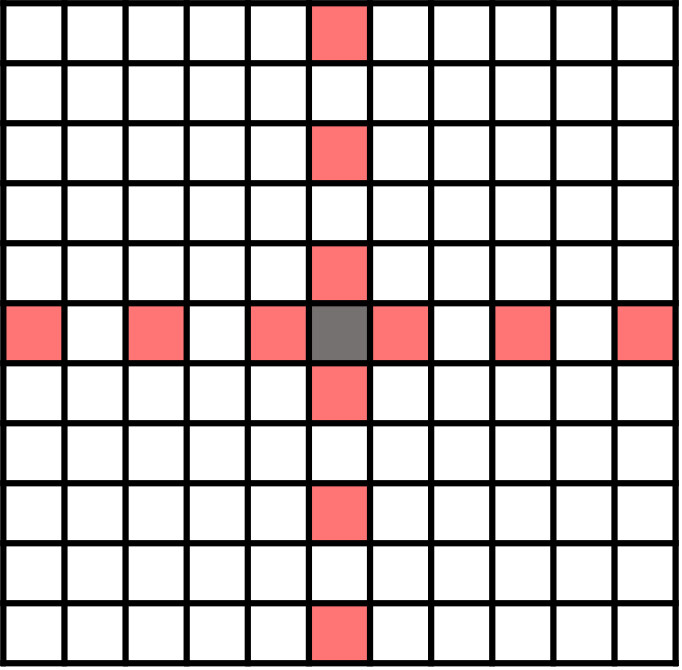}
        \\
        (a) & (b) & (c)
    \end{tabular}
    \caption{
        \textbf{Red-Black PatchMatch Propagation Kernels.} 
        In each kernel, the grey square indicates the pixel to be updated. 
        The red squares indicate the neighboring pixels that provide PatchMatch a set of candidate oriented points for the gray pixel. We use kernel (c) for the coarsest level and kernel (b) for the finer levels.
    }
    \label{fig:red-black-kernel} 
 \end{figure}

The oriented point map $\omega^t$ at the $t$-th PatchMatch iteration is propagated according to the propagation kernel. A common kernel is the Red-Black propagation kernel by Galliani \etal~\cite{galliani2015massively}, as illustrated in Figure \ref{fig:red-black-kernel}. We let $\psi^{t}_{:, p} = \{\omega^t_{q} \mid q \in K(p) \} \cup \{{\omega^t_p}^{prt} \}$ denote the set of candidate oriented points obtained by propagation kernel $K$ at pixel $p$ and by random perturbation of the current candidate. The propagation can be applied using a series of convolutional filters of one-hot encodings, with one values in positions that correspond to each neighbor, as defined by $K$.  The visibility-weighted feature correlations for each candidate are computed as $\mathcal{G}^{\mathcal{V}}_{\psi_{k, p}} = \frac{\sum_{v \in \mathcal{V}_p} \hat{\mathcal{V}}^{v}_p\mathcal{G}^v_{\psi_{k,p}} }{ \sum_{v\in \mathcal{V}_p} \hat{\mathcal{V}}^v_p}$.

\subsection{Candidate Regularized Cost and Update}
\label{sec:recurrent_cost_regularization}

\begin{figure}[t]
    \centering
    \includegraphics[width=0.47\textwidth]{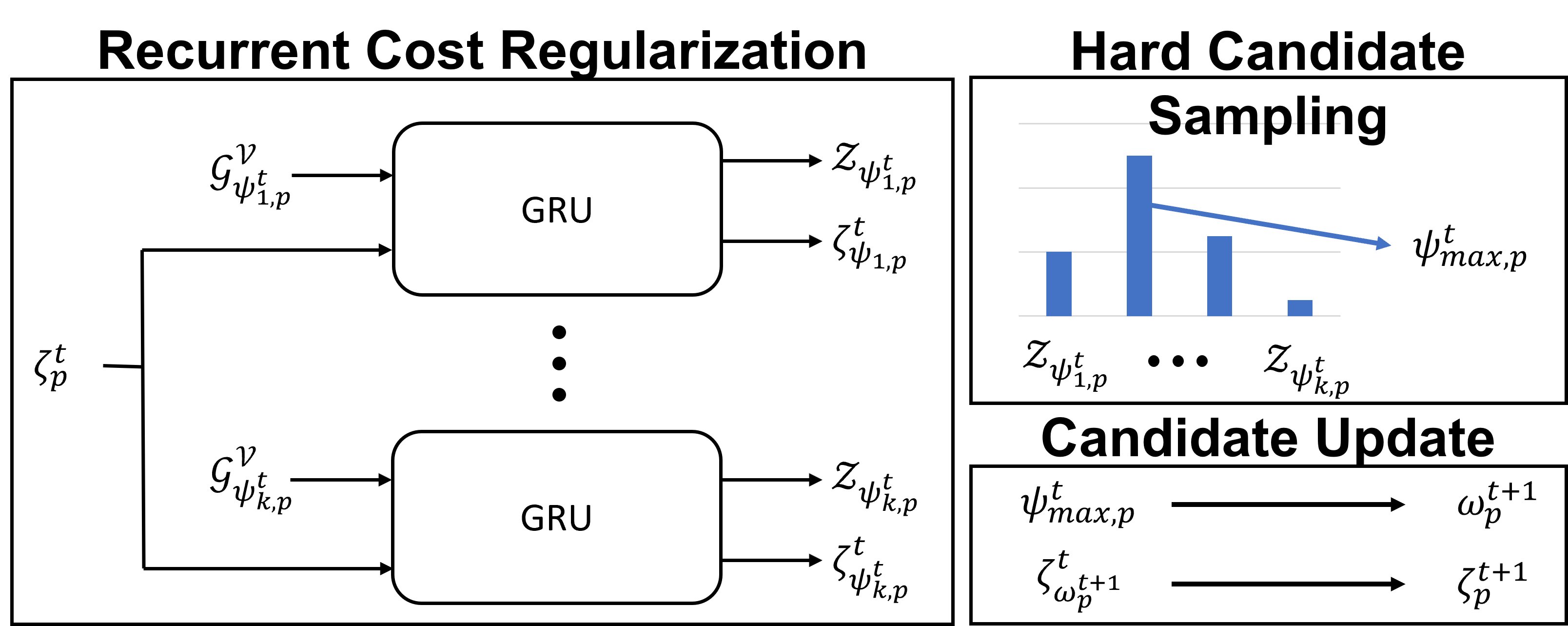}
    \caption{
        \textbf{Recurrent Cost Regularization.}
        Given the hidden state $\zeta_p^t$ of each pixel $p$ in the $t$-th iteration and the visibility-weighted feature correlations of each propagated candidate $\mathcal{G}^{\mathcal{V}}_{\psi^t_{:, p}}$, we use a Gated Recurrent Unit (GRU) module to estimate the regularized cost $\mathcal{Z}_{\psi_{:, p}}$ and updated hidden state $\zeta^{t}_{\psi_{:, p}}$ for each plane candidate. 
        Then, the best candidate $\psi_{max, p}$ for the next iteration is hard-sampled according to the regularized costs, replacing the current oriented point $\omega^{t}_p$ at $p$, and the corresponding hidden states of the pixel $\zeta^{t}_p$ are updated using the corresponding sampled candidate $\zeta^{t}_{\omega^{t+1}_p}$.
    }
    \label{fig:recurrent_cost_regularization}
 \end{figure}

Existing learning-based cost regularization methods, such as 3D convolution on spatially aligned cost volume~\cite{yao2018mvsnet} or $k$-nearest neighbor based graph convolutions~\cite{chen2019point}, exploit ordinal relationships between neighboring label maps. However, there is no consistent relationship between candidates for $\omega_p$ or for candidates of neighboring pixels.
Instead, we get insight from loopy Belief-Propagation (LBP), where each node's belief is iteratively updated by message-passing from the neighboring nodes, so that confidently labeled nodes propagate to less confident neighbors.  We represent beliefs for each candidate as hidden states $\zeta^t_{\psi_{k, p}}$, and use a recurrent neural network (RNN) to estimate regularized score $\mathcal{Z}_{\psi_{k, p}}$ and updated hidden state $\zeta^{t+1}_{\psi_{k, p}}$.
Figure~\ref{fig:recurrent_cost_regularization} illustrates the process. 

Similar to LBP, we compute pairwise neighborhood smoothness~\cite{besse2012pmbp} of the candidate with respect to the current label, $\{ M(\psi_{k, p}, \omega_q) | q \in \mathcal{N}(p) \}$, where $M(\omega_p, \omega_q) = dist(\omega_p, q) + dist(\omega_q, p)$ is the sum of distances between each oriented point and the plane parameterized by the other oriented point. 
We append the smoothness terms to the weighted feature correlation $\mathcal{G}^{\mathcal{V}}_{\psi_{k, p}}$ as an input to the RNN. The RNN can then aggregate the confidences (represented by feature correlations) over similar oriented points. 

The per-pixel candidates and corresponding hidden states are updated by:
\begin{align*}
    \omega^{t+1}_p &= \psi^t_{k, p} \sim \mathcal{Z}^t_{\psi_{:, p}}\\
    \zeta^{t+1}_p &= \zeta^t_{\omega^{t+1}_{p}}.
\end{align*}
In inference, the sampling of $\omega_p$ is $\arg\max$; in training, the sampling hardens from probabilistic to
$\arg\max$ as training progresses. The updated hidden states are used as an input to the recurrent cost regularization module in the next PatchMatch iteration.

\subsection{Coarse-to-Fine PatchMatch and Fusion}

The estimated map of oriented points $\omega^t$ and the corresponding hidden states $\zeta^t$ are upsampled as an input to the finer level PatchMatch iteration using nearest neighbor interpolation. 
The $\omega$ of the finest level are fused together into a 3D point cloud by following the method used by other MVS systems~\cite{galliani2015massively, schoenberger2016mvs,yao2018mvsnet}. First, consistency is checked for each reference image with the source views using reprojection distance, relative depth distance, and normal consistency. Then, we reproject the mean value of $N$-view consistent depths into the world space to obtain consensus points. 
\begin{figure*}[t]
    \centering
    \begin{tabular}{@{}c@{}cc@{}cc@{}c@{}}
        \includegraphics[width=0.16\textwidth]{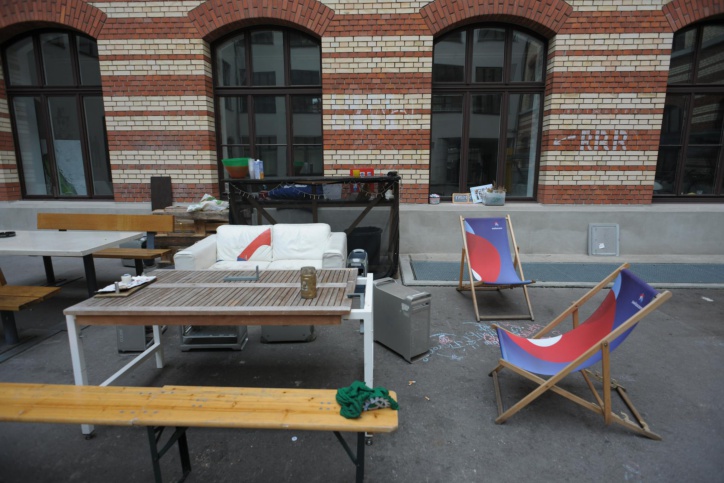} &
        \includegraphics[width=0.16\textwidth]{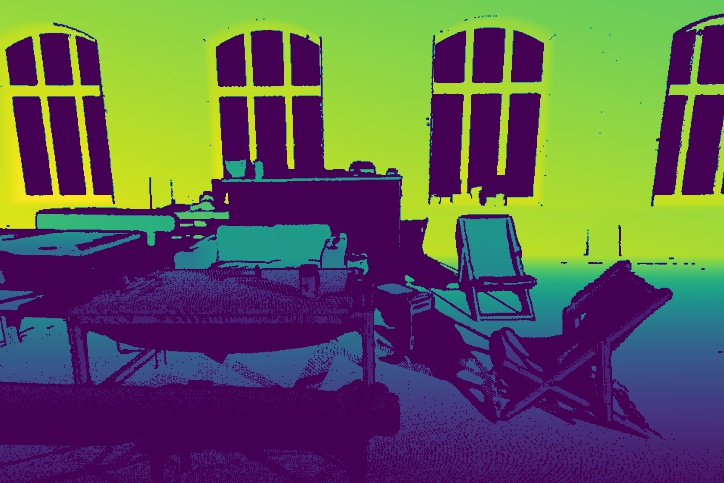} &
        \includegraphics[width=0.16\textwidth]{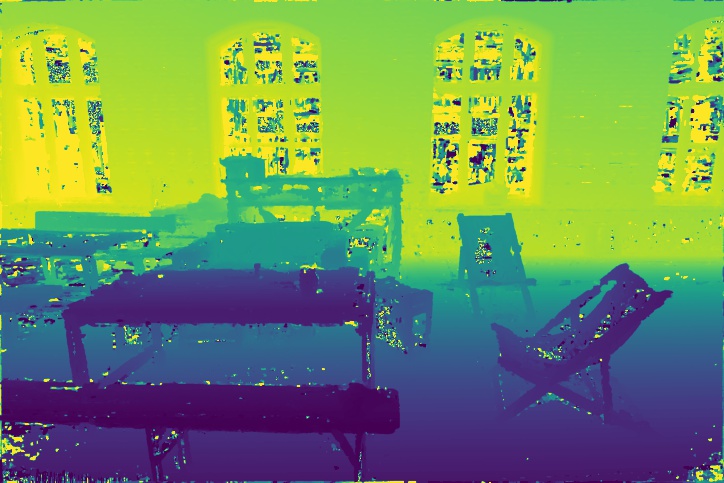} &
        \includegraphics[width=0.16\textwidth]{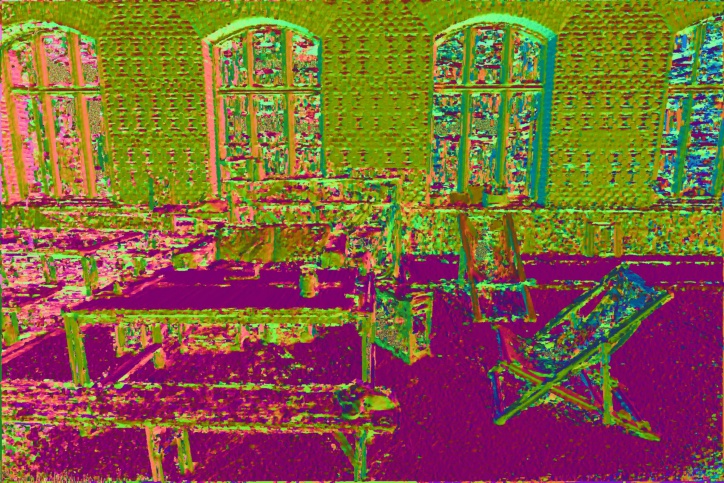} &
        \includegraphics[width=0.16\textwidth]{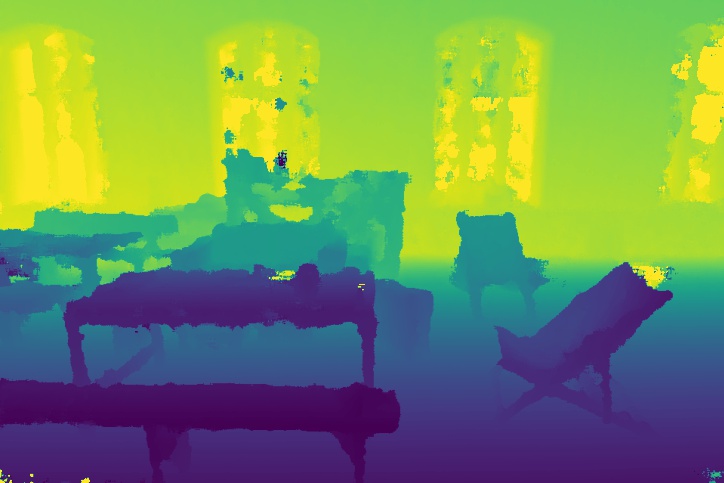} &
        \includegraphics[width=0.16\textwidth]{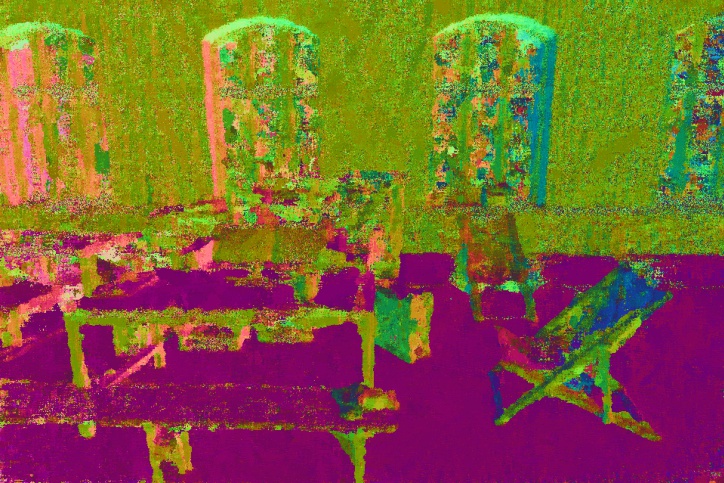}
        \\
        \includegraphics[width=0.16\textwidth]{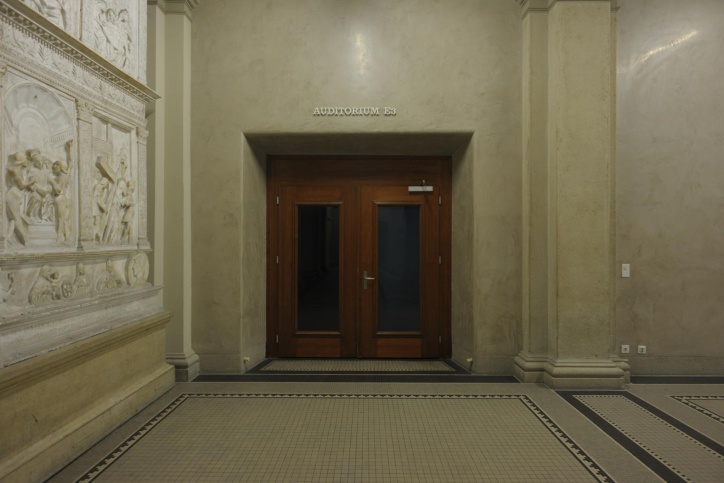} &
        \includegraphics[width=0.16\textwidth]{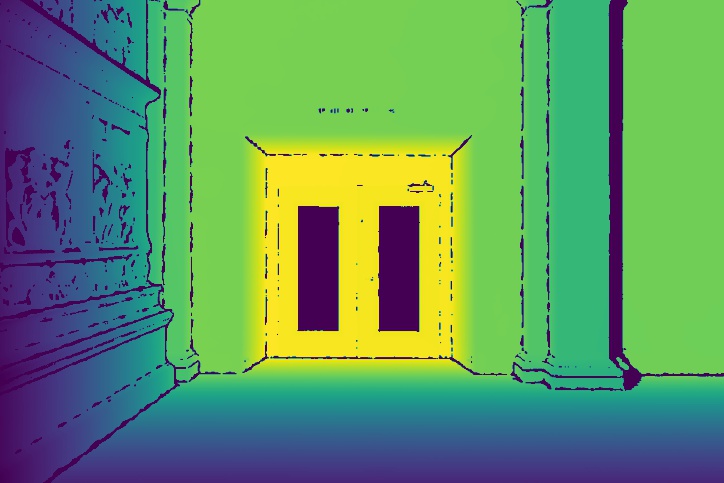} &
        \includegraphics[width=0.16\textwidth]{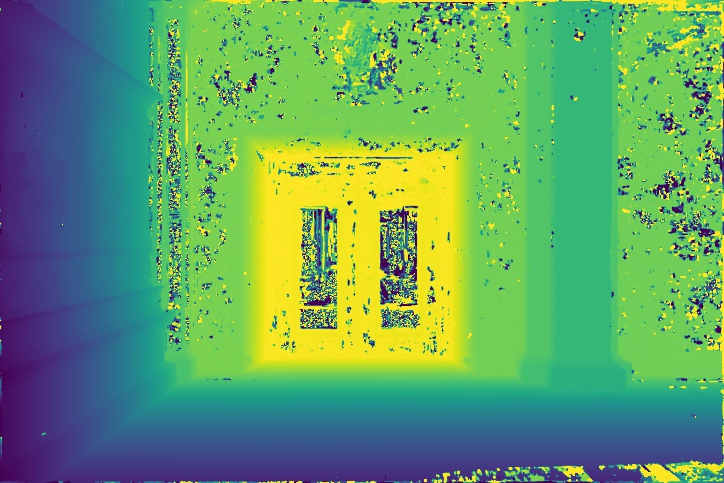} &
        \includegraphics[width=0.16\textwidth]{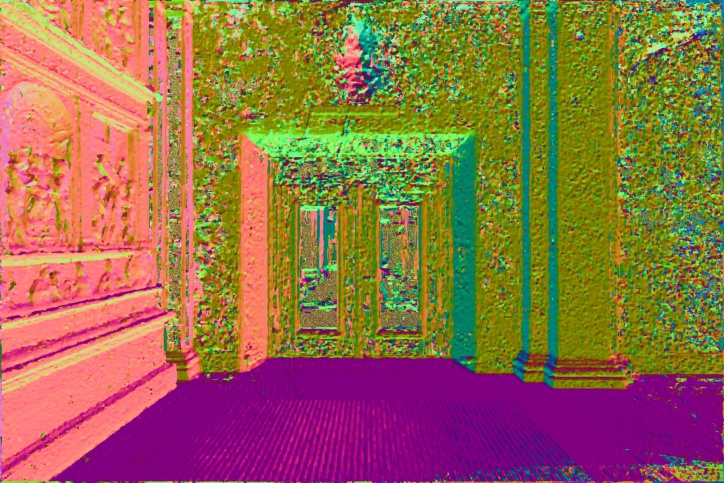} &
        \includegraphics[width=0.16\textwidth]{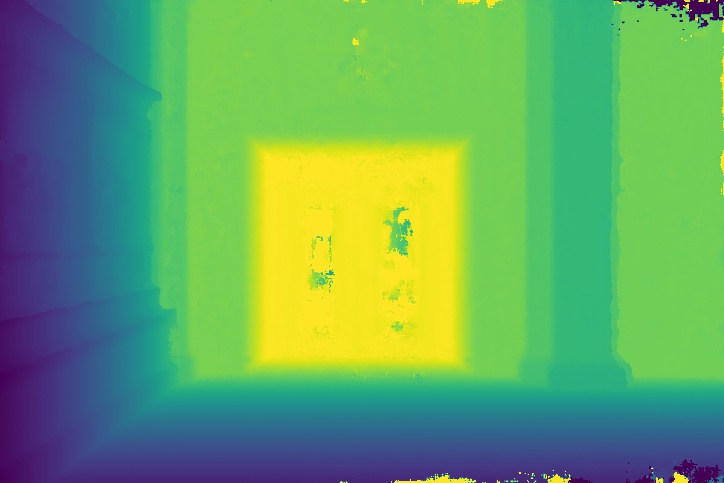} &
        \includegraphics[width=0.16\textwidth]{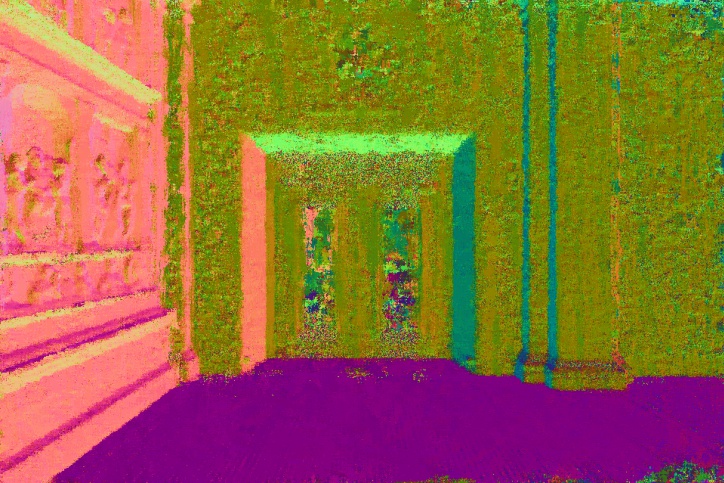}
        \\
        \includegraphics[width=0.16\textwidth]{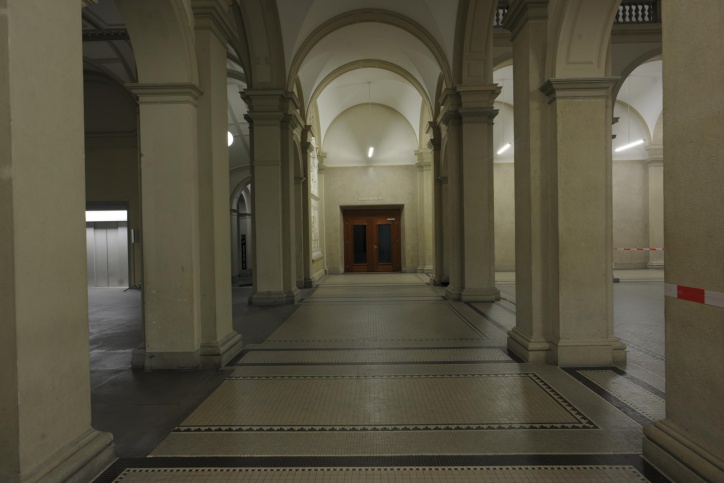} &
        \includegraphics[width=0.16\textwidth]{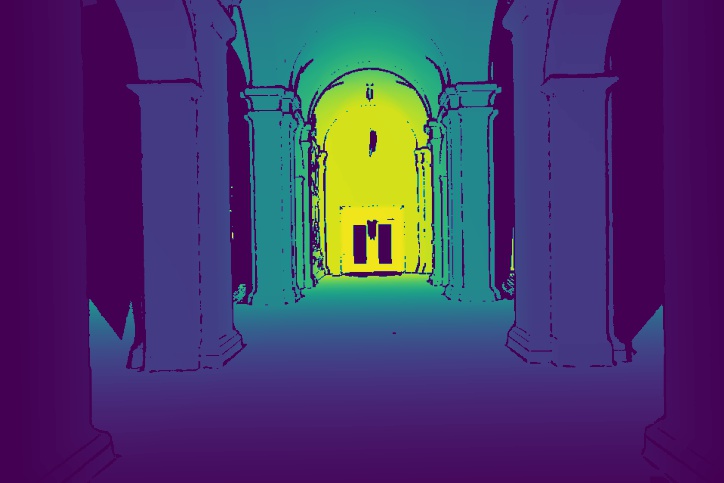} &
        \includegraphics[width=0.16\textwidth]{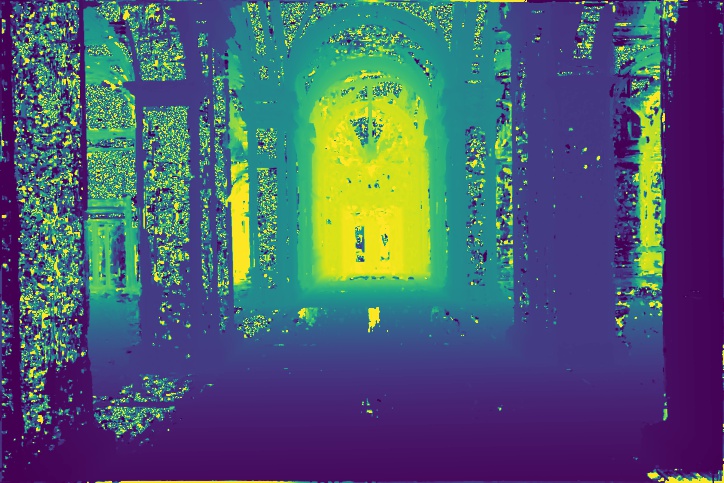} &
        \includegraphics[width=0.16\textwidth]{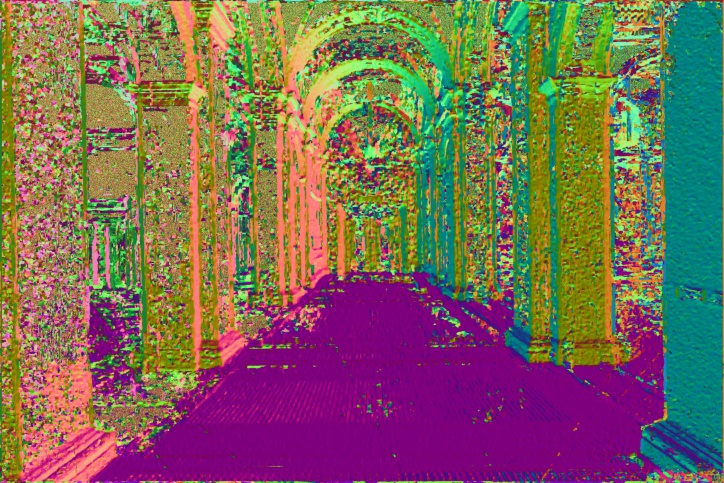} &
        \includegraphics[width=0.16\textwidth]{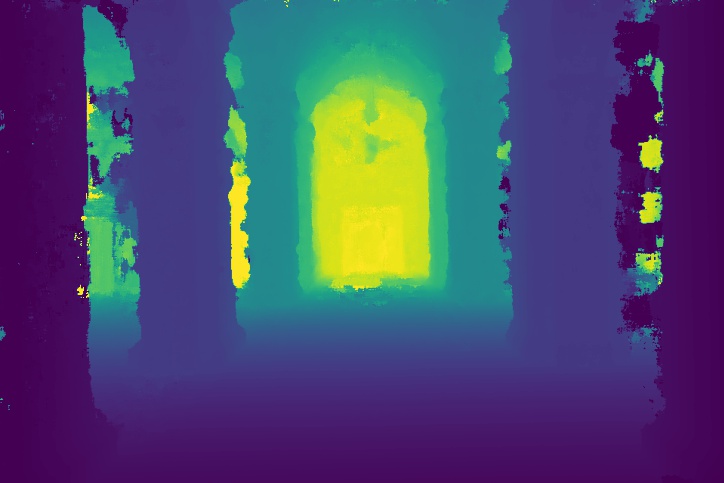} &
        \includegraphics[width=0.16\textwidth]{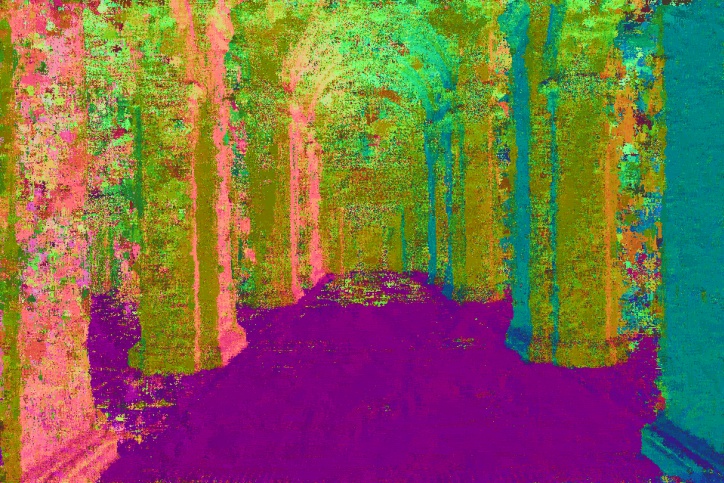}
        \\
        \includegraphics[width=0.16\textwidth]{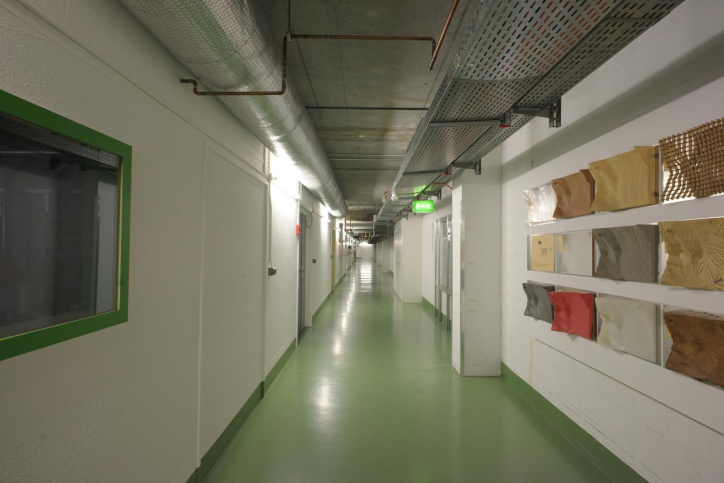} &
        \includegraphics[width=0.16\textwidth]{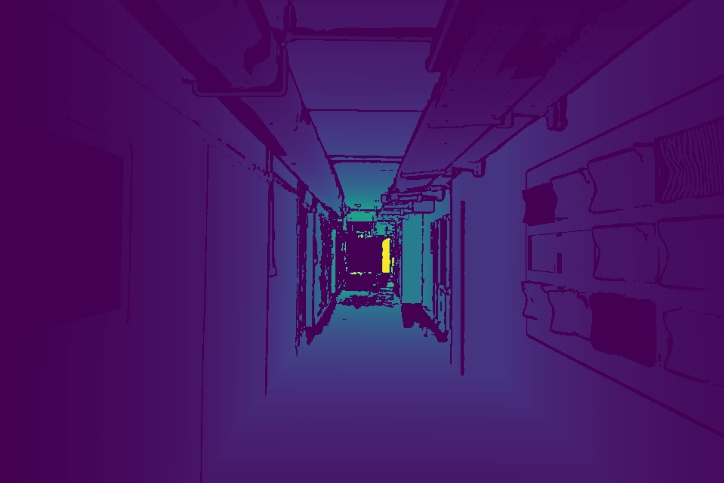} &
        \includegraphics[width=0.16\textwidth]{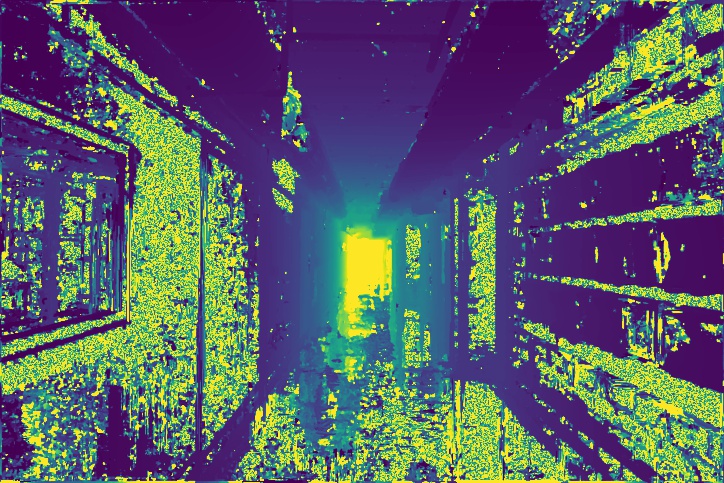} &
        \includegraphics[width=0.16\textwidth]{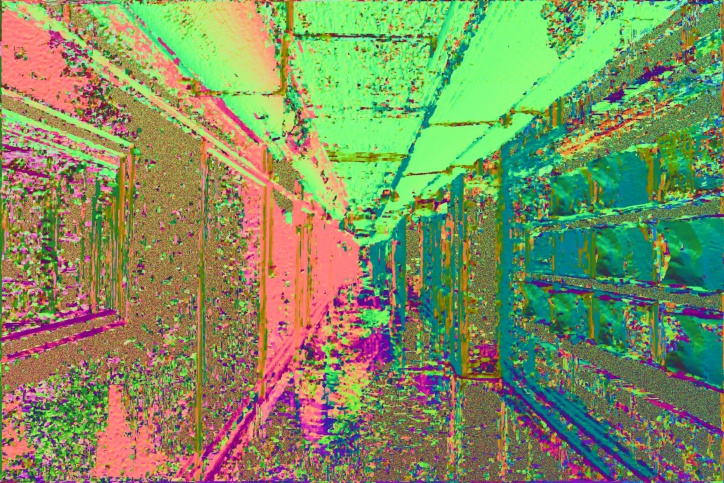} &
        \includegraphics[width=0.16\textwidth]{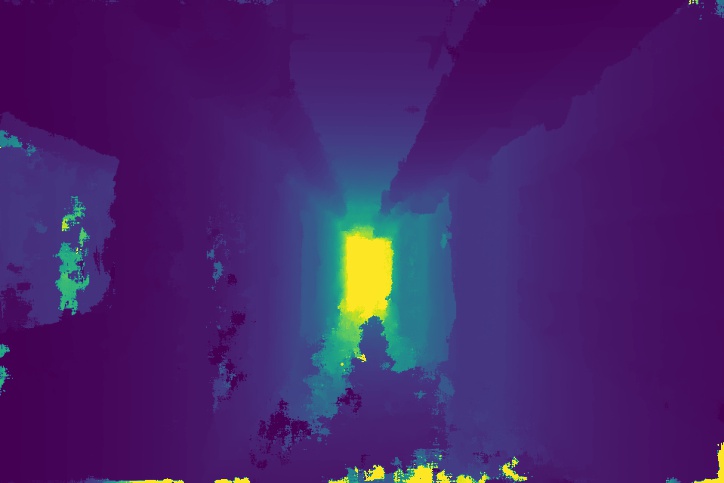} &
        \includegraphics[width=0.16\textwidth]{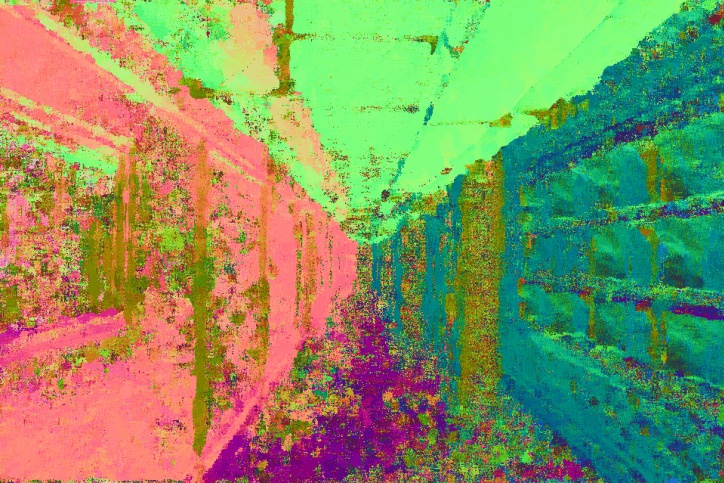}
        \\
        Ref. Image & GT. Depth & \multicolumn{2}{c}{COLMAP} & \multicolumn{2}{c}{Ours}
    \end{tabular}
    \caption{
        \textbf{Qualitative comparison against COLMAP on the ETH3D high-resolution benchmark.} 
        From the left, reference image, ground truth depth, depth estimate from COLMAP, normal estimate from COLMAP, depth estimate of our model, and normal estimate of our model. All of the depth maps share the same color scale based on the ground truth depth ranges. We show that our estimated depths and normals are more complete than COLMAP.
    }
    \vspace{-0.1in}
    \label{fig:eth3d} 
\end{figure*}

\section{PatchMatch-RL Training}

It is challenging to make PatchMatch MVS end-to-end trainable.  The $\arg\max$ based hard decisions/sampling required for PatchMatch update and view selection is non-differentiable, and the incorporation of normal estimates with soft-$\arg\max$ causes depth and normal to depend on each other. 
We propose a reinforcement learning approach to jointly learn the candidate cost and visibility estimation parameters.

We use $\mathcal{V}^{\theta_\mathcal{V}}(\omega_p)$ to denote the pixel-wise visibility estimation function, parameterized by $\theta_{\mathcal{V}}$, that outputs visibility score $\hat{\mathcal{V}}^s_p$ for each source image $s$ given images $\mathcal{I}$ and cameras $\mathcal{C}$. We use  $\mathcal{S}^{\theta_\mathcal{S}}(\psi_p)$ to denote a matching score function, parameterized by $\theta_\mathcal{S}$, that produces plane candidate score $\mathcal{Z}_{\psi_p}$ for each $\psi_p$ given $\mathcal{I}, \mathcal{C}$ and selected views $\mathcal{V}_p$.
Our formulation contains two agents: one selects views and the other selects the candidates. 

\subsection{Reward Function}
We define the reward $r^t = \mathcal{N}(\omega^t; \omega^*, \sigma_\omega)$ as a probability of observing the oriented point $\omega^t$ from distribution given ground truth oriented point value $\omega^*$ in iteration $t$. We define the distribution as a joint independent normal distribution of depth and normal of pixel $p$:
\begin{equation}
    \mathcal{N}(\omega^t_p; \omega^*_p, \sigma_\omega) = \mathcal{N}(n^t_p; n^*_p, \sigma_n) \cdot \mathcal{N}(d^t_p; d^*_p, \sigma_d).
\end{equation}
We let the expected future reward be a $\gamma$-discounted sum of future rewards:  $G^t = \sum_{t' >= t} \gamma^{t'- t} r^t$. 
We formulate the gradient of the reward as a negation of the gradient of cross-entropy between the step-wise reward $\mathcal{N}(\omega^t; \omega^*, \sigma_\omega)$ and an agent $\pi_\theta(a^t, s^t)$, according to the REINFORCE algorithm as:
\begin{align}
    \triangledown_\theta J &= \mathrm{E}_{\pi_\theta}[Q^{\pi_\theta}(s, a) \triangledown_\theta \log \pi_\theta(a \mid s)]\nonumber\\
    &= \sum_t \triangledown_\theta \ln \pi^t_\theta(a^t \mid s^t) G^t\nonumber\\
    &= \sum_t \sum_{t'>=t} \triangledown_\theta \gamma^{t'-t}(\mathcal{N}(\omega^{t'}; \omega^*, \sigma_\omega) \log \pi_\theta).
\end{align}

The sampling can be done in two ways: the categorical distribution, which makes the policy approximate the expectation of the distribution; or argmax, which makes the policy the greedy solution. As an exploration versus exploitation strategy, we employ a decaying $\epsilon$-greedy approach where we sample candidates using (1) expectation by probability of $\epsilon$ or (2) using argmax by probability of $1 - \epsilon$. We also apply a decaying reward of $\tau \cdot \mathcal{N}(d^t_p; d^*_p, \sigma_d)$.

Below, we describe the policy of each agent. We use $\mathbb{S}_\mathcal{V}, \mathbb{A}_\mathcal{V}, \pi_\mathcal{V}, \mathbb{R}_\mathcal{V}$, and $\mathbb{S}_\mathcal{S}, \mathbb{A}_\mathcal{S}, \pi_\mathcal{S}, \mathbb{R}_\mathcal{S}$ to denote the state, action, policy and reward space of the view selection and candidate selection agents respectively. For simplicity, we use $s^t \in \mathbb{S}$, $a^t \in \mathbb{A}$, and $r^t \in \mathbb{R}$ to denote the corresponding agent's state, action, and reward in the $t$-th iteration that apply to a particular pixel.

\subsection{Learning Photometric Cost}
For the candidate selecting agent, the state space is the set of candidate plane parameters $\psi_{:}$ for each oriented point $\omega_p$, and the the action space is the selection of a candidate label for each pixel in each iteration according to the parameterized photometric cost function $\mathcal{S}^{\theta_\mathcal{S}}(\omega_p)$. 
The probability of selecting each candidate is defined as a softmax distribution based on the photometric cost of each plane candidate, and 
the stochastic policy 
$\pi_\mathcal{S}$ samples from this distribution:

\begin{align}
    \pi_\mathcal{S}(a^t \mid s^t) &= 
    \omega^t \sim \frac{e^{-\mathcal{S}^{\theta_\mathcal{S}}(\psi^t_:)}}{\sum_{q \in K} e^{-\mathcal{S}^{\theta_\mathcal{S}}(\psi^t_q)}}
    \label{eq:policy_candidate}
\end{align}
The parameters can be learned via gradient ascent through the negative cross-entropy between the probability distribution of the candidates given ground truth and the probability distribution of the candidates estimated by photometric cost function: 

\begin{align*}
    \triangledown_{\theta_\mathcal{S}} \mathcal{N}_{\omega^{t}} \log \pi_\mathcal{S}
    &= \triangledown_{\theta_\mathcal{S}} \sum_{k\in K} \mathcal{N}_{\psi^t_k} \cdot
        \log(
            \frac{
                e^{-\mathcal{S}^{\theta_\mathcal{S}(\psi^t_k)}}
            }{
                \sum_{j \in K}
                e^{-\mathcal{S}^{\theta_\mathcal{S}}(\psi^t_{j})}
            }
        )\nonumber
\end{align*}
where $\mathcal{N}_{\psi^t_k} = \mathcal{N}(\psi^t_k; \omega^*, \sigma_\psi)$ represents the probability of observing the candidate $\psi^t_k$ according to the ground truth.

\subsection{Learning View Selection}
For the view selection agent, the state space contains the set of source images; the action space is a selection of $N$ images among the source images for each iteration; and the policy uses the parameterized view selection function $\mathcal{V}(\omega^t_p)$ to estimate the visibility $(\forall s \in I_{src}),  \hat{\mathcal{V}^s}$. 
The stochastic policy $\pi_\mathcal{V}$ is:
\begin{align}
    \label{eq:policy_view}
    \pi_\mathcal{V}(a^t \mid s^t) &= v \sim \frac{
            \hat{\mathcal{V}}^v
        } {
            \sum_{s \in I_{src}} \hat{\mathcal{V}}^s
        }
\end{align}
and the gradient:
\begin{align}
    \triangledown_{\theta_\mathcal{V}} \log \pi_\mathcal{V}
    &= \triangledown_{\theta_\mathcal{V}} \log(
        \frac{\sum_{v\in N}\hat{\mathcal{V}}^v}
             {\sum_{s \in I_{src}} \hat{\mathcal{V}}^s})\nonumber \\
    \label{eq:policy_gradient_view}
    &\approx \triangledown_{\theta_\mathcal{V}}(
        \log(\sum_{v\in N}\hat{\mathcal{V}}^v) - 
        \log(\sum_{m \in (N \cup M)} \hat{\mathcal{V}}^m))\nonumber .
\end{align}
For robustness of training, we include only the selected $N$ views and worse $M$ views in the denominator to prevent minimizing the probabilities of good but unselected views.  This incentivizes training to assign more visibility to good views than bad views (that do not view the point corresponding to the reference pixel).


\begin{table*}[t]
\begin{small}
\resizebox{\textwidth}{!}{
\begin{tabular}{llccccccc}
 &  & & \multicolumn{3}{c}{\cellcolor[HTML]{D9EAD3}\textbf{Test 2cm: Accuracy / Completeness / F1}} & \multicolumn{3}{c}{\cellcolor[HTML]{D9EAD3}\textbf{Test 5cm: Accuracy / Completeness / F1}} \\
\multicolumn{1}{l}{\textbf{Method}} & \multicolumn{1}{c}{\textbf{Resolution}} & \multicolumn{1}{c}{\textbf{Time(s)}} & \cellcolor[HTML]{D9EAD3}\textbf{Indoor} & \cellcolor[HTML]{D9EAD3}\textbf{Outdoor} & \cellcolor[HTML]{D9EAD3}\textbf{Combined} & \cellcolor[HTML]{D9EAD3}\textbf{Indoor} & \cellcolor[HTML]{D9EAD3}\textbf{Outdoor} & \cellcolor[HTML]{D9EAD3}\textbf{Combined} \\
\midrule
ACMH~\cite{xu2019multi}             & 3200x2130 & 546.77 & \textbf{91.1 / 64.8 / 73.9} & \textbf{84.0 / 80.0 / 81.8} & \textbf{89.3 / 68.6 / 75.9} & 97.4 / 78.0 / 83.7 & 94.1 / 75.0 / 90.4 & 96.6 / 87.1 / 85.4 \\
Gipuma~\cite{galliani2015massively} & 2000x1332 & 272.81 & 86.3 / 31.4 / 41.9 & 78.8 / 45.3 / 55.2 & 84.4 / 34.9 / 45.2 & 95.8 / 42.1 / 54.9 & 93.8 / 54.3 / 67.2 & 95.3 / 45.1 / 58.0 \\
COLMAP~\cite{schoenberger2016mvs}   & 3200x2130 & 2245.57 & 92.0 / 59.7 / 70.4 & 92.0 / 73.0 / 80.8 & 92.0 / 63.0 / 73.0 & 96.6 / 73.0 / 82.0 & 97.1 / 83.9 / 89.7 & 96.8 / 75.7 / 84.0 \\
\midrule
PVSNet~\cite{xu2020pvsnet}          & 1920x1280 & -      & 65.6 / 78.6 / 70.9 & 68.8 / 84.3 / 75.7 & 66.4 / 80.1 / 72.1 & 82.4 / 87.8 / 84.7 & 84.5 / 92.7 / 88.2 & 82.9 / 89.0 / 85.6 \\
PatchmatchNet~\cite{wang2020patchmatchnet}
                                    & 2688x1792 & 491.69 & 68.8 / 74.6 / 71.3 & 72.3 / 86.0 / 78.5 & 69.7 / 77.5 / 73.1 & 84.6 / 85.1 / 84.7 & 87.0 / 92.0 / 89.3 & 85.2 / 86.8 / 85.9 \\
Ours                                & 1920x1280 & 556.50 & 73.2 / 70.0 / 70.9 & 78.3 / 78.3 / 76.8 & 74.5 / 72.1 / 72.4 & \textbf{88.0 / 83.7 / 85.5} & \textbf{92.6 / 89.0 / 90.5} & \textbf{89.2 / 85.0 / 86.8} \\
 &  & & \multicolumn{1}{l}{} & \multicolumn{1}{l}{} & \multicolumn{1}{l}{} & \multicolumn{1}{l}{} & \multicolumn{1}{l}{} & \multicolumn{1}{l}{} \\
 &  & & \multicolumn{3}{c}{\cellcolor[HTML]{CFE2F3}\textbf{Train 2cm: Accuracy / Completeness / F1}} & \multicolumn{3}{c}{\cellcolor[HTML]{CFE2F3}\textbf{Train 5cm: Accuracy / Completeness / F1}} \\
\multicolumn{1}{l}{\textbf{Method}} & \multicolumn{1}{c}{\textbf{Resolution}} &\multicolumn{1}{c}{\textbf{Time(s)}} &  \cellcolor[HTML]{CFE2F3}\textbf{Indoor} & \cellcolor[HTML]{CFE2F3}\textbf{Outdoor} & \cellcolor[HTML]{CFE2F3}\textbf{Combined} & \cellcolor[HTML]{CFE2F3}\textbf{Indoor} & \cellcolor[HTML]{CFE2F3}\textbf{Outdoor} & \cellcolor[HTML]{CFE2F3}\textbf{Combined} \\
\midrule
ACMH~\cite{xu2019multi}             & 3200x2130 & 486.35 & \textbf{92.6 / 59.2 / 70.0} & \textbf{84.7 / 64.4 / 71.5} & \textbf{88.9 / 61.6 / 70.7} & 97.7 / 70.1 / 80.5 & 95.4 / 75.6 / 83.5 & 96.6 / 72.7 / 81.9 \\
Gipuma~\cite{galliani2015massively} & 2000x1332 & 243.34 & 89.3 / 24.6 / 35.8 & 83.2 / 25.3 / 37.1 & 86.5 / 24.9 / 36.4 & 96.2 / 34.0 / 47.1 & 95.5 / 36.7 / 51.7 & 95.9 / 35.2 / 49.2 \\
COLMAP~\cite{schoenberger2016mvs}   & 3200x2130 & 2102.71 & 95.0 / 52.9 / 66.8 & 88.2 / 57.7 / 68.7 & 91.9 / 55.1 / 67.7 & 98.0 / 66.6 / 78.5 & 96.1 / 73.8 / 82.9 & 97.1 / 69.9 / 80.5 \\
\midrule
PatchmatchNet~\cite{wang2020patchmatchnet}
                                    & 2688x1792 & 473.92 & 63.7 / 67.7 / 64.7 & 66.1 / 62.8 / 63.7 & 64.8 / 65.4 / 64.2 & 78.7 / 80.0 / 78.9 & 86.8 / 73.2 / 78.5 & 82.4 / 76.9 / 78.7 \\
Ours                                & 1920x1280 & 555.58 & 76.6 / 60.7 / 66.7 & 75.4 / 64.0 / 69.1 & 76.1 / 62.2 / 67.8 & \textbf{89.6 / 76.5 / 81.4} & \textbf{88.8 / 81.4 / 85.7} & \textbf{90.5 / 78.8 / 83.3}
\end{tabular}}
\end{small}
\caption{
\label{tab:eth3d}
\textbf{Results on the ETH3D high-resolution MVS benchmark train and test sets.} We do not train on any ETH3D data. Bold denotes the method with the highest F1 score for each setting. Results from several other methods are shown for comparison. We measure the mean time taken for reconstructing each scene (including the fusion stage) using the author provided code on the same hardware.  
PVSNet results are not available on the train set. Our method outperforms other recent learning-based approaches (PVSNet and PatchmatchNet) in most of the metrics. 
} 
\end{table*}

\begin{table}[t]
\begin{small}
\centering
    \begin{tabular}{llc}
                                        & \multicolumn{2}{c}{\cellcolor[HTML]{CFE2F3}\textbf{Precision / Recall / F1}}                                  \\
    \multicolumn{1}{c}{\textbf{Method}} & \multicolumn{1}{c}{\cellcolor[HTML]{CFE2F3}\textbf{Intermediate}} & \cellcolor[HTML]{CFE2F3}\textbf{Advanced} \\
    CIDER~\cite{xu2020learning}                & 42.8 / 55.2 / 46.8                                             & 26.6 / 21.3 / 23.1                     \\
    COLMAP~\cite{schoenberger2016mvs}          & 43.2 / 44.5 / 42.1                                             & 33.7 / 24.0 / 27.2                     \\
    R-MVSNet~\cite{yao2019recurrent}           & 43.7 / 57.6 / 48.4                                               & 31.5 / 22.1 / 24.9                     \\
    CasMVSNet~\cite{gu2020cascade}             & 47.6 / 74.0 / 56.8                                             & 29.7 / 35.2 / 31.1                     \\
    AttMVS~\cite{luo2020attention}             & \textbf{61.9 / 58.9 / 60.1}                                    & 40.6 / 27.3 / 31.9                     \\
    PatchmatchNet~\cite{wang2020patchmatchnet} & 43.6 / 69.4 / 53.2                                             & 27.3 / 41.7 / 32.3                     \\
    PVSNet~\cite{xu2020pvsnet}                 & 53.7 / 63.9 / 56.9                                             & \textbf{29.4 / 41.2 / 33.5}            \\
    BP-MVSNet~\cite{sormann2020bpmvsnet}       & 51.3 / 68.8 / 57.6                                              & 29.6 / 35.6 / 31.4                     \\
    Ours                                       & 45.9 / 62.3 / 51.8                                              & 30.6 / 36.7 / 31.8                    
    \end{tabular}
\end{small}
    \caption{
        \textbf{Results on the Tanks and Temples benchmark.} 
        The best performing model based on $F_1$ score is marked as bold. 
        Similar to Table~\ref{tab:eth3d}. Our method performs on par with existing learning based methods on the advanced sets.
    }
    \label{tab:tnt}
\end{table}

\section{Experiments}
We evaluate our work on two large-scale benchmarks: Tanks and Temples Benchmark~\cite{Knapitsch2017tanks} and ETH3D High-Res Multi-View Benchmark~\cite{schoeps2017eth3d}.

\subsection{Training Details}
For all experiments, we train using the BlendedMVS dataset~\cite{yao2020blendedmvs}, which contains a combination of 113 object, indoor, and outdoor scenes with large viewpoint variations. We use the low-res version of the dataset which has a spatial resolution of $768 \times 576$.  Throughout training and evaluation, we use $\alpha=3$ and $\beta=3$, 3 layers of hidden states $\mathcal{H}$, $\gamma = 0, 1$ for photometric scorer and view selection scorer respecively, and feature map sizes corresponding to $\frac{1}{8}, \frac{1}{4}$ and $\frac{1}{2}$ of the original image size.
For training, we use $2$, $1$, and $1$ iterations, and for evaluation we use $8$, $2$, and $2$ iterations for each scale respectively. 
We use the PatchMatch Kernel $K$ shown in Figure~\ref{fig:red-black-kernel}(b) for training. 
As an exploitation versus exploration strategy, we employ a Decaying $\epsilon$-Greedy approach where we either sample candidates proportional to their softmax scores with a probability of $\epsilon$ or select the argmax candidate with the probability of 1 - $\epsilon$. The initial value of $\epsilon$ is 0.9 with an exponential decay of 0.999 per each step.

To promote view-selection robustness, for each reference image, we select 6 total views from the same scene: 3 random views and 3 views sampled from the 10 best views according to BlendedMVS. 
Among 6 source images, we sample 1 best visibility-scoring image as visible and 2 worst visibility-scoring images as invisible. 
We train the model with Adam~\cite{kingma2015adam} and set the initial learning rate to $0.001$ and the decay to $0.5$ per epoch. 
We implemented our approach in PyTorch. We use an Nvidia RTX 3090 for training and evaluation. 

\subsection{ETH3D High-Res Multi-View Benchmark}
We evaluate our method on the ETH3D High-res Multi-View Benchmark, which contains  17 different indoor or outdoor scenes with 6048x4032 resolution images for each scene.
For evaluation, we fix the number of the source views to 10 and sample the 3 best views. 
We use a fixed image resolution of $1920 \times 1280$ with camera intrinsics obtained by COLMAP~\cite{schoenberger2016mvs}. 
The system takes 13.5 seconds and uses 7693MB of peak memory for each reference image.
Table~\ref{tab:eth3d} provides quantitative results. 
We show that our method achieves comparable results to the other listed methods on the standard 2cm benchmark and the best results on the 5cm benchmark. Most learning-based methods fail to produce reasonable results on ETH3D because there are few images with wide baselines and large depth ranges. 
In Figure~\ref{fig:eth3d}, we compare the inferred depth and normal maps with COLMAP~\cite{schoenberger2016mvs}. 
From the results, we can see that our method can cover weakly textured regions, such as white walls and pillars, more completely than COLMAP~\cite{schoenberger2016mvs}, while still maintaining good accuracy. 
However, the model may fail to reconstruct reflective surfaces and large texture-less areas. 

\begin{figure*}[t]
    \centering
    \includegraphics[width=0.48\textwidth]{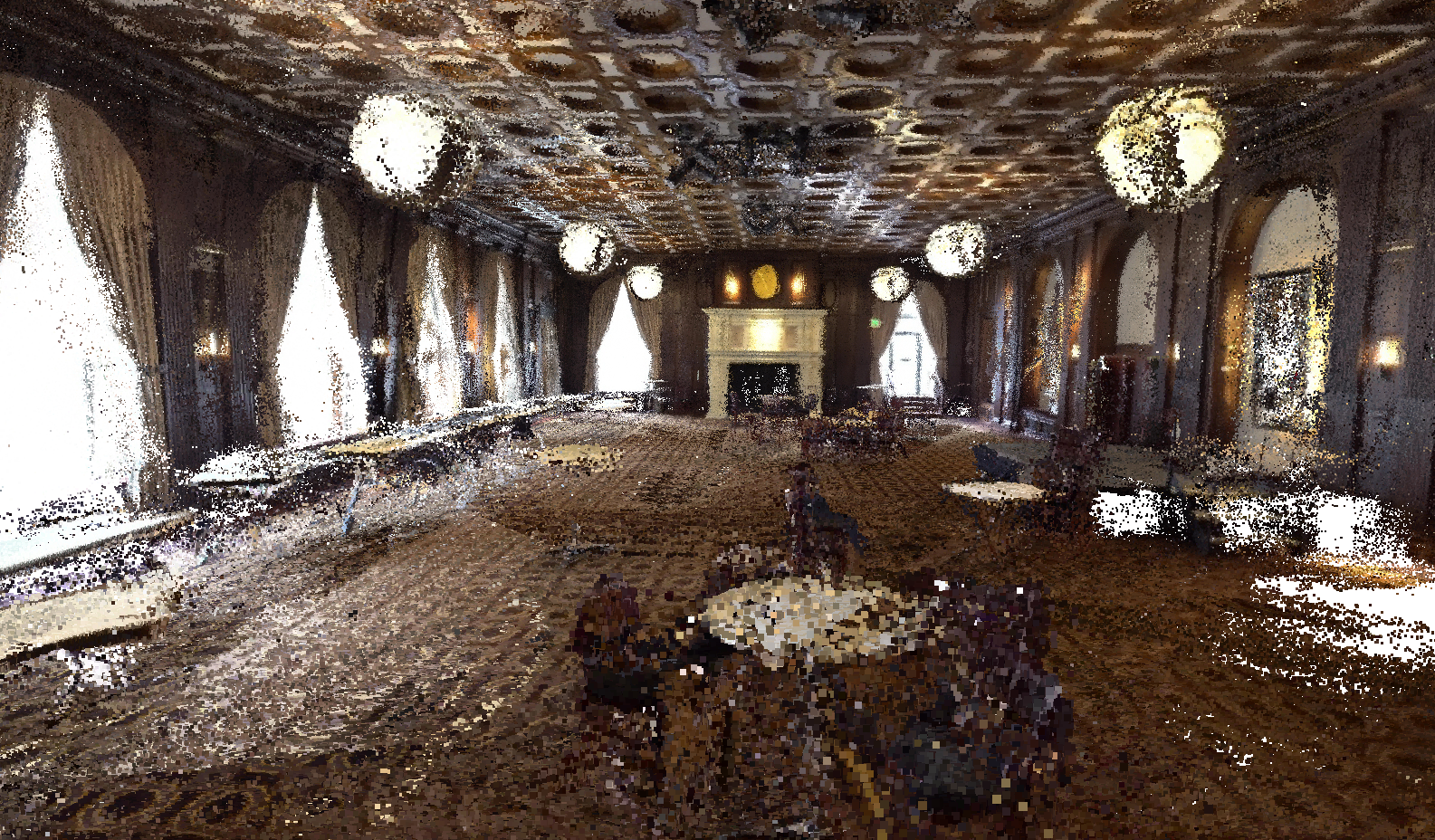}
    \includegraphics[width=0.48\textwidth]{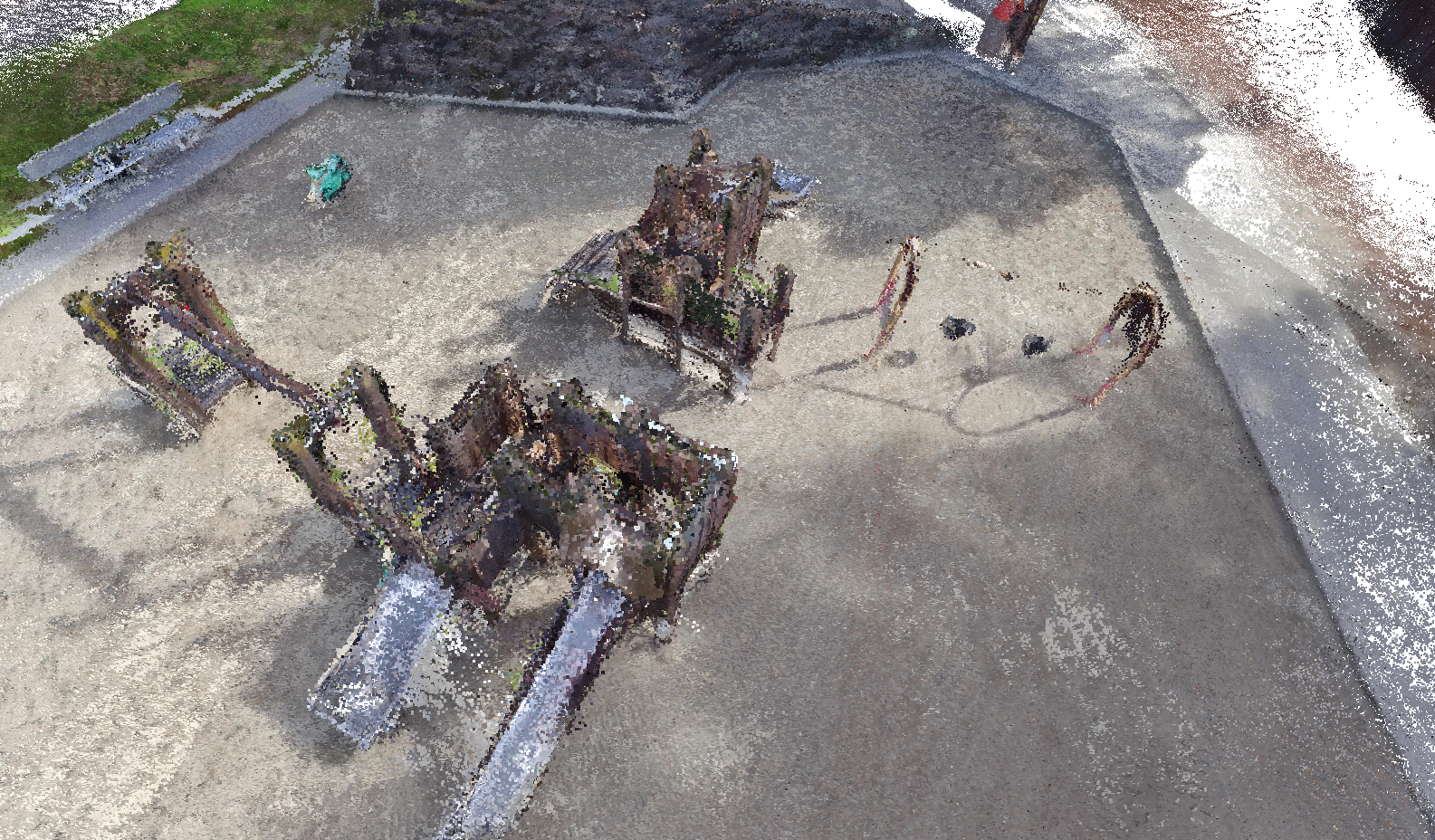} \\
    \includegraphics[width=0.325\textwidth]{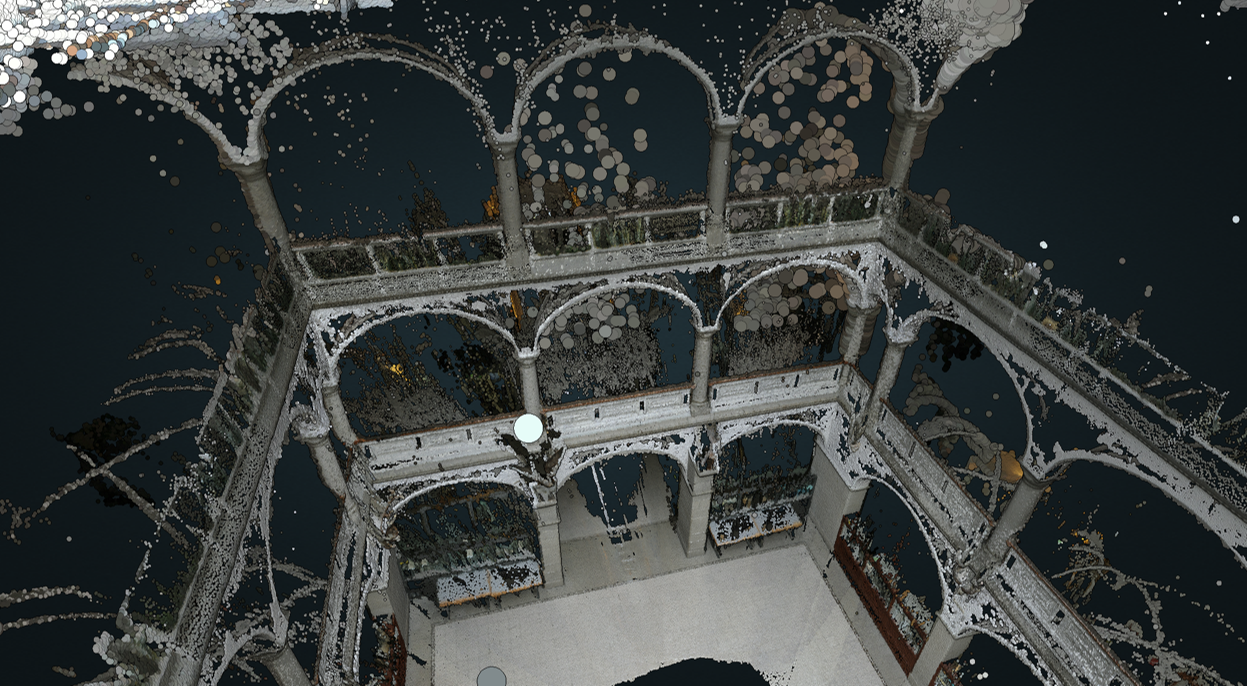}
    \includegraphics[width=0.325\textwidth]{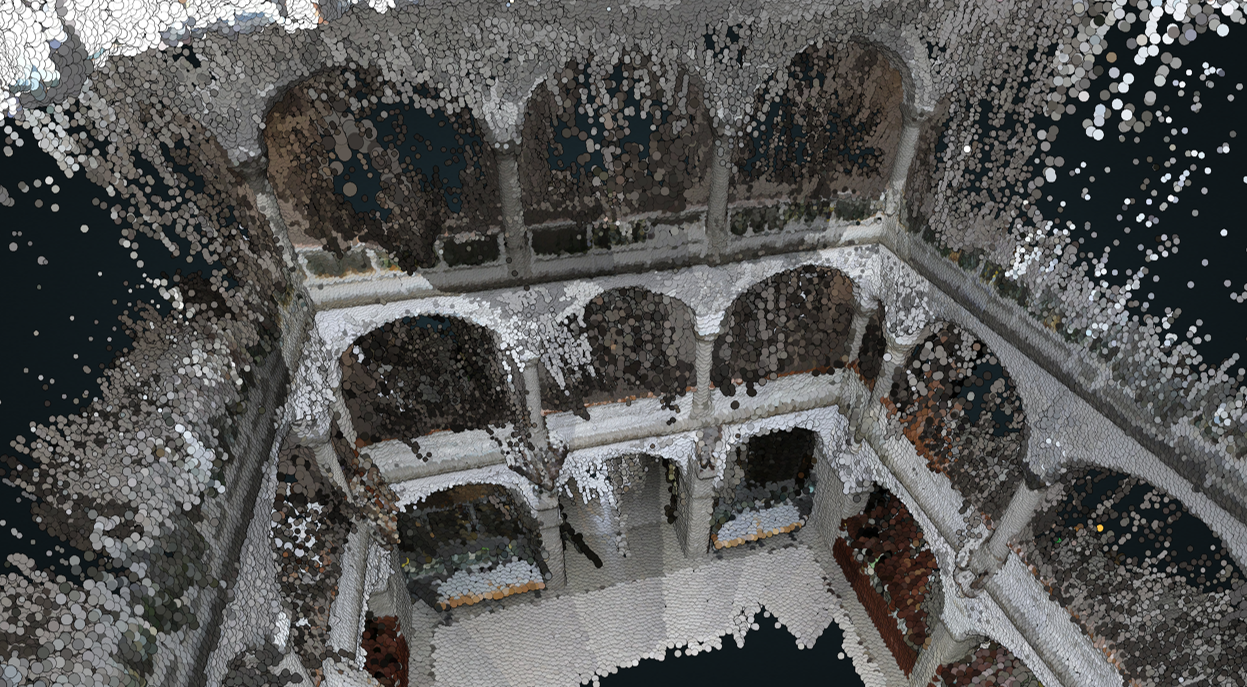}
    \includegraphics[width=0.325\textwidth]{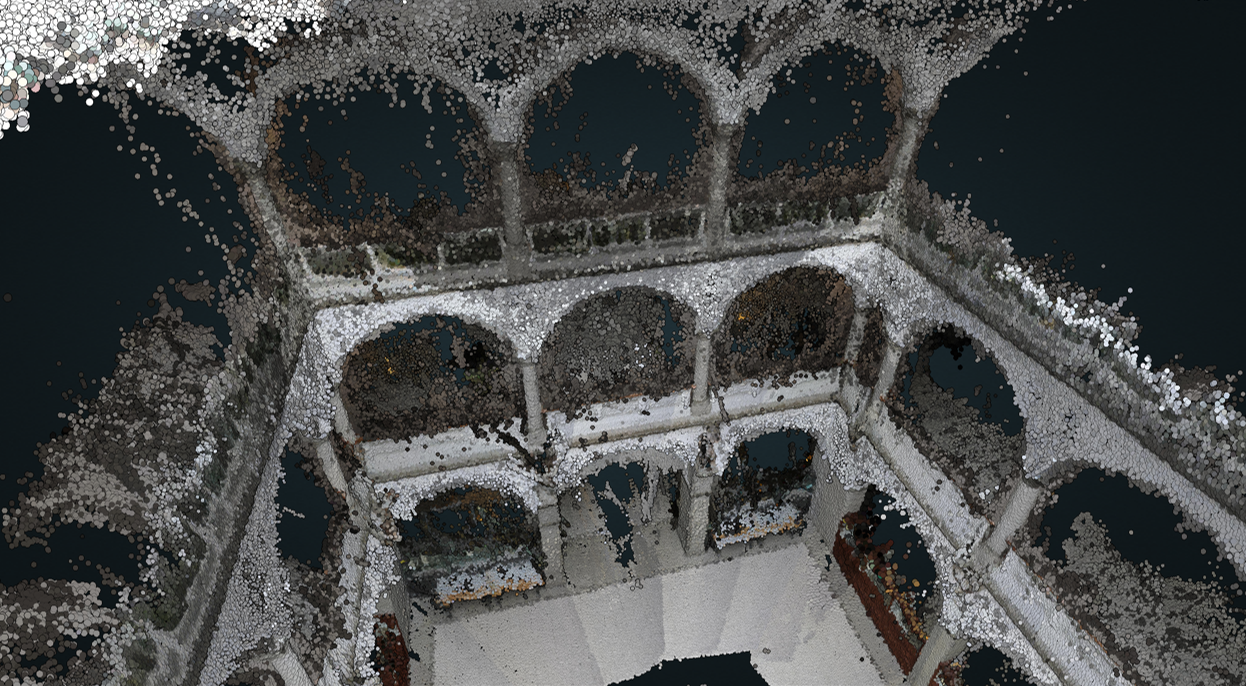}
    \caption{
        \textbf{Point cloud reconstruction results.}
        For the top row, from left to right, we show \textit{Ballroom} and \textit{Playground} from the Tanks and Temples benchmark~\cite{Knapitsch2017tanks}.
        For the bottom row, from left to right, we show the reconstruction results of COLMAP~\cite{schoenberger2016mvs}, PatchmatchNet~\cite{wang2020patchmatchnet} and our method in the \textit{Exhibition Hall} from the ETH3D benchmark~\cite{schoeps2017eth3d}. See the \href{https://www.eth3d.net/result_details?id=289}{benchmark website} for more reconstruction results.
    }
        \vspace{-0.1in}
    \label{fig:pointcloud} 
\end{figure*}
 
\begin{table}[t]
\begin{small}
    \begin{tabular}{lcc}
    \multicolumn{1}{l}{} & \multicolumn{2}{c}{\cellcolor[HTML]{CFE2F3}\textbf{Accuracy / Completeness / F1}}       \\
    \textbf{Model}    & \cellcolor[HTML]{CFE2F3}\textbf{Train 2CM} & \cellcolor[HTML]{CFE2F3}\textbf{Train 5cm} \\
    w/o normal               & 62.9 / 58.0 / 54.0                      & 81.1 / 76.7 / 75.1                      \\
    w/o view sel.            & 75.8 / 56.7 / 64.1                      & 89.4 / 72.9 / 79.8                      \\
    w/o rcr.                 & 75.6 / 60.9 / 66.7                      & 89.0 / 77.9 / 82.6                      \\
    Ours                 & \textbf{76.1 / 62.2 / 67.8}             & \textbf{90.5 / 78.8 / 83.3}            
    \end{tabular}
    \caption{
        \textbf{Ablation Study on ETH3D High-Res Training Set.} We compare our original system to using 1x1 feature patches that do not take advantage of normal estimates (``w/o normal''), without pixelwise view selection, instead using the top 3-ranked source images for all pixels, and without recurrent cost regularization (``w/o rcr'').   
        The model with the highest $F_1$ score is marked with bold for each threshold.
    }
    \label{tab:ablation}
\end{small}
\end{table}

\subsection{Tanks and Temples Benchmark}
With the same trained model, we evaluate on the Tanks and Temples~\cite{Knapitsch2017tanks} intermediate and advanced benchmarks which contain 
8 intermediate and 6 advanced large-scale scenes respectively. 
Similar to the ETH3D High-res benchmark, we fix the number of the source views to 10, sample the 3 best views, and fix the image resolution to $1920 \times 1080$.
Our method takes 12.1 seconds and uses 5801MB of peak memory for each reference image.
Table~\ref{tab:tnt} shows the quantitative results of the benchmark. 
We achieve similar results to CasMVSNet~\cite{gu2020cascade} and PatchmatchNet~\cite{wang2020patchmatchnet}.
In Figure~\ref{fig:pointcloud}, we present qualitative results on the reconstructed point clouds. 
We show that our method can generate complete and accurate reconstructions, which includes repeated textures such as carpet and thin structures such as poles for the swing. 

\subsection{Ablation Studies}
Table~\ref{tab:ablation} shows how each component contributes to the performance of our method. 

\noindent\textbf{Importance of normals}:
Our use of normals enables modeling oblique surfaces and provides a locally planar support region for photometric costs, which has otherwise been achived through deformable convolution~\cite{wang2020patchmatchnet} or $k$-NN~\cite{chen2019point}. 
Without normal estimation for more non-frontal planar propagation and support, accuracy drops by 13.2\% and completeness drops by 3.2\% for the 2cm threshold on ETH3D (Table~\ref{tab:ablation} ``w/o normal'').

\noindent\textbf{Importance of pixelwise view selection}:
Without pixelwise selection of the source images, the completeness of the reconstruction at the 2cm threshold drops by 5.5\% and accuracy drops slightly (Table~\ref{tab:ablation} ``w/o view sel''). Pixelwise view selection makes better use of many source views that are partially overlapping the reference image. 

\noindent\textbf{Importance of recurrent cost regularization}:
We introduce recurrent cost regularization to aggregate confidences (i.e. feature correlations) across similar points without requiring aligned cost volumes. For comparison, we try scoring candidates using a multi-layer network based on only the feature correlations for the centered patch. With this simplification, the overall $F1$ score drops by 1.1\% for the 2cm threshold (Table~\ref{tab:ablation} ``w/o rcr ''). 

\noindent\textbf{Importance of argmax sampling}:
We tried to train using soft-argmax based candidates where we take the expectation of the normals and the depths independently. However, we failed to train an effective model due to the initial aggregated values being clustered to the middle of the depth ranges, which limits the range of predictions. Existing works may avoid this problem by using sampling in a more restrictive way; e.g., by performing a depth search with reduced range~\cite{duggal2019deeppruner} or by sampling initial points from uniformly separated bins~\cite{wang2020patchmatchnet}. Our reinforcement learning approach enables us to perform argmax sampling in the same way as non-learning based approaches while benefiting from learned representations.


\vspace{-1.5mm}
\section{Conclusion}
\vspace{-1.5mm}

We propose an end-to-end trainable MVS system that estimates pixelwise depths, normals, and visibilities using PatchMatch optimization.  We use reinforcement learning and a decaying $\epsilon$-greedy sampling in training to learn effectively despite using view selection and argmax sampling in inference. Our system performs well compared to the latest learning-based MVS systems, but further improvements are possible.  For example, we have not yet incorporated some of the sophisticated geometric checks of ACMM~\cite{xu2019multi} or post-process refinement of DeepC-MVS~\cite{kuhn2020deepc}, and higher resolution processing would also yield better results. By incorporating most of the critical ideas from non-learning based methods into a learning-based framework, our work provides a promising direction for further improvements in end-to-end
approaches.


\vspace{-1.5mm}
\section*{Acknowledgements}
\vspace{-1.5mm}
\noindent We thank ONR MURI Award N00014-16-1-2007 for support in our research.

{\small
\bibliographystyle{ieee_fullname}
\bibliography{egbib}
}

\end{document}